\documentclass[letterpaper]{article} 
\usepackage{aaai23}  
\usepackage{times}  
\usepackage{helvet}  
\usepackage{courier}  
\usepackage[hyphens]{url}  
\usepackage{graphicx} 
\usepackage{natbib}  
\usepackage{caption} 
\usepackage{multirow}
\usepackage[ruled,noline,noalgohanging]{algorithm2e}
\usepackage{stmaryrd}
\usepackage{tikz}
\usetikzlibrary{arrows,positioning,decorations.markings}
\tikzset{
  varnode/.style={rectangle,outer sep=0mm},
  varnodenoperi/.style={rectangle,outer sep=-1mm},
  ourarrow/.style={>=stealth}, 
  ourarc/.style={>=stealth,thick,arc}}
\usepackage{amsthm}
\usepackage{amssymb}

\newtheorem{mydef}{Definition}
\newtheorem{myenc}{Encoding}
\newtheorem{mydec}{Decoding}
\newtheorem{mythm}{Theorem}

\newtheorem{myeg}{Example}

\usepackage{multicol}
\usepackage{latexsym}
\usepackage[inline]{enumitem}
\setlist[enumerate]{label=(\roman*)}
\usepackage[Euler]{upgreek}
\usepackage{tabularx}
\usepackage{mathtools,amsmath}
\usepackage{subcaption}
\usepackage{isabelle, isabellesym}

\title{Formally Verified SAT-Based AI Planning}
\author{Mohammad Abdulaziz$^{1,2}$\ \ \ \ \ Friedrich Kurz$^2$}
\affiliations{1. King's College London, The United Kingdom\\ 2. Techniche Universit\"at M\"unchen, Germany}

\usepackage{textcomp}
\usepackage{listings}
\lstdefinestyle{inline}{
    basicstyle=\ttfamily\small
}

\usepackage{bm}

\newcommand{\fred}[1]{Fred: \textcolor{cyan}{#1}}
\newcommand{\todo}[1]{\bgroup\color{red}#1\egroup}
\usepackage[draft]{hyperref}
\usepackage{graphicx}
\newcommand{\spacex}[2]{#2}
\usepackage[normalem]{ulem}
\usepackage{lineno}

\usepackage[most]{tcolorbox}
\usepackage{stmaryrd}
\tcbuselibrary{breakable,hooks,skins,theorems}
\newtcbtheorem{IsabelleTheorem}{Listing}{
    enhanced jigsaw
    , breakable
    , float=h
    , top=0pt
    , right=0pt
    , bottom=0pt
    , left=0pt
    , boxrule=0pt
    , toprule=1pt
    , bottomrule=1pt
    , titlerule=1pt
    , colframe=black
    , sharp corners
    , colbacktitle=white
    , coltitle=black
    , fonttitle=\tiny
    , colback=white
    , fontupper=\ttfamily\tiny
    , before upper={\parindent0em}
}{isa}

\lstdefinelanguage{isabelle}{
    morekeywords={record,type_synonym,definition,fun,function,primrec,where,lemma,theorem,unfolding,by,shows,assumes,and,datatype,using,abbreviation
,moreover,have,hence,thus,qed,proof,let,ultimately,show,next,in}
    , sensitive=true
    , showstringspaces=false
    , framerule=0pt
    , xleftmargin=2em
    , numbers=left
    , numberstyle=\ttfamily\tiny
    , firstnumber=1
    , stepnumber=2
    , basicstyle=\ttfamily\tiny
    , breaklines=true
    , showspaces=false
    , morecomment=[l]{--}
    , morecomment=[s]{(*}{*)}
    , commentstyle=\color{gray}
    , morestring=[b]"
    , literate={\\<times>}{{$\times$}}{1} {\\<equiv>}{{$\equiv$}}{1} {\\<forall>}{{$\forall$}}{1} {\\<exists>}{{$\exists$}}{1} {\\<and>}{{$\land$}}{1}
        {\\<in>}{{$\in$}}{1} {\\<Rightarrow>}{{$\Rightarrow$}}{1} {\\<lambda>}{{$\lambda$}}{1} {::}{{$::$}}{1}
        {\\<subseteq>}{{$\subseteq$}}{1} {\\<^sub>m}{{$_m$}}{1} {\\<longleftrightarrow>}{{$\longleftrightarrow$}}{3}
        {\\<pi>}{{$\pi$}}{1} {\\<delta>}{{$\delta$}}{1} {\\<lbrakk>}{{$\llbracket$}}{1} {\\<rbrakk>}{{$\rrbracket$}}{1}
        {\\<Longrightarrow>}{{$\Longrightarrow$}}{3} {\\<not>}{{$\lnot$}}{1} {\\<le>}{{$\le$}}{1} {\\<rightharpoonup>}{{$\rightharpoonup$}}{2}
        {\\<^sub>\\<V>}{{$_{\mathcal V}$}}{1} {\\<lparr>}{{$\llparenthesis$}}{1} {\\<rparr>}{{$\rrparenthesis$}}{1}
        {\\<leftarrow>}{{$\leftarrow$}}{1} {\\<^sub>\\<O>}{{$_{\mathcal O}$}}{1} {\\<^sub>I}{{$_{\texttt{I}}$}}{1}
        {\\<^sub>G}{{$_{\texttt{G}}$}}{2} {\\<phi>}{{$\varphi$}}{1} {\\<Phi>}{{$\Phi$}}{1} {\\<psi>}{{$\psi$}}{1} {\\<Psi>}{{$\Psi$}}{1}
        {\\<^sub>S}{{$_{\texttt S}$}}{1} {\\<inverse>}{{$^{-1}$}}{1} {\\<^sub>O}{{$_{\texttt O}$}}{1} {\\<^bold>\\<And>}{{$\bm\bigwedge$}}{1}
        {\\<^bold>\\<or>}{{$\bm\lor$}}{1} {\\<^sub>G}{{$_{\texttt G}$}}{1} {\\<Pi>}{{$\Pi$}}{1} {\\<^sub>I}{{$_{\texttt I}$}}{1} {\\<noteq>}{{$\neq$}}{1}
        {\\<bottom>}{{$\bot$}}{1} {\\<^sub>+}{{$_\texttt +$}}{1} {\\<^bold>\\<and>}{{$\bm\land$}}{1} {\\<^bold>\\<not>}{{$\bm\lnot$}}{1}
        {\\<^sub>1}{{$_1$}}{1} {\\<^sub>2}{{$_2$}}{1} {\\<A>}{{$\mathcal A$}}{1} {\\<Turnstile>}{{$\models$}}{2} {\\<^sub>\\<forall>}{{$_\forall$}}{1}
        {\\<^sub>0}{{$_0$}}{1} {\\<tau>}{{$\tau$}}{1}  {\\<^sub>\\<Omega>}{{$_\Omega$}}{1} {\\<^sub>V}{{$_V$}}{1} {\\<^bold>\\<Or>}{{$\bm\bigvee$}}{1}
        {\\<^sub>P}{{$_\texttt P$}}{1} {\\<^sub>X}{{$_\texttt X$}}{1} {\\<longrightarrow>}{{$\longrightarrow$}}{2} {\\<or>}{{$\lor$}}{1}
}

\newtcblisting[auto counter]{IsabelleSnippet}[2][]{
    listing options={
        language=isabelle
    }
    , listing only
    , enhanced jigsaw
    , breakable
    , top=-.5em
    , right=0pt
    , bottom=-.5em
    , left=0pt
    , boxrule=0pt
    , toprule=1pt
    , bottomrule=1pt
    , titlerule=1pt
    , colframe=black
    , sharp corners
    , colbacktitle=white
    , coltitle=black
    , fonttitle=\tiny
    , colback=white
    , fontupper=\ttfamily\tiny
    , before upper={\parindent0em}
    , title=Listing \thetcbcounter: #2
    , #1
}

\usepackage[htt]{hyphenat}

\newcommand{\fullversion}[1]{#1}
\begin{document}
\maketitle

\renewcommand{\rangle}{)}
\renewcommand{\langle}{(}
\begin{abstract}
We present an executable formally verified SAT encoding of ground classical AI planning problems.
We use the theorem prover Isabelle/HOL to perform the verification.
We experimentally test the verified encoding and show that it can be used for reasonably sized standard planning benchmarks.
We also use it as a reference to test a state-of-the-art SAT-based planner, showing that it sometimes falsely claims that problems have no solutions of certain lengths.

\end{abstract}

\providecommand{\insts}{}
\renewcommand{\insts}{\ensuremath{\Delta}}
\providecommand{\inst}{\ensuremath{\tvsal}}
\newcommand{\act}{\ensuremath{\pi}}
\newcommand{\asarrow}[1]{\vec{#1}}
\renewcommand{\vec}[1]{\overset{\rightarrow}{#1}}
\newcommand{\as}{\ensuremath{\vec{{\act}}}}

\newcommand{\etc}{\textit{etc.}}
\newcommand{\versus}{\textit{vs.}}
\newcommand{\ie}{i.e.}
\newcommand{\Ie}{I.e.}
\newcommand{\eg}{e.g.}
\newcommand{\michael}[1]{\textcolor{blue}{M: #1}}
\newcommand{\abziz}[1]{\textcolor{brown}{#1}}
\newcommand{\sublist}[2]{ \ensuremath{#1} \preceq\!\!\!\raisebox{.4mm}{\ensuremath{\cdot}}\; \ensuremath{#2}}
\newcommand{\subscriptsublist}[2]{\ensuremath{#1}\preceq\!\raisebox{.05mm}{\ensuremath{\cdot}}\ensuremath{#2}}
\newcommand{\PLS}{\Pi^\preceq\!\raisebox{1mm}{\ensuremath{\cdot}}}
\newcommand{\PLScharles}{\Pi^d}
\newcommand{\execname}{\mathsf{ex}}
\newcommand{\IndHyp}{\mathsf{IH}}
\newcommand{\exec}[2]{#2(#1)}

\newcommand{\ancestorssymbol}{\textsf{\upshape ancestors}}
\newcommand{\ancestors}{\ancestorssymbol}
\newcommand{\satpreas}[2]{\ensuremath{sat_precond_as(s, \as)}}
\newcommand{\proj}[2]{\ensuremath{#1{\downharpoonright}_{#2}}}
\newcommand{\dep}[3]{\ensuremath{#2 {\rightarrow} #3}}
\newcommand{\deptc}[3]{\ensuremath{#2 {\rightarrow^+} #3}}
\newcommand{\negdep}[3]{\ensuremath{#2 \not\rightarrow #3}}
\newcommand{\leavessymbol}{\textsf{\upshape leaves}}
\newcommand{\leaves}{\leavessymbol}

\newcommand{\childrensymbol}{\textsf{\upshape children}}
\newcommand{\children}[2]{\mathcal{\childrensymbol}_{#2}(#1)}
\newcommand{\succsymbol}{\textsf{\upshape succ}}
\newcommand{\succstates}[2]{\succsymbol(#1, #2)}
\newcommand{\concat}{\#}
\newcommand{\RG}{\cite{Rintanen:Gretton:2013}\ }
\newcommand{\KG}{Kovacs' grammar}
\newcommand{\cupdot}{\charfusion[\mathbin]{\cup}{\cdot}}
\newcommand{\cuparrow}{\charfusion[\mathbin]{\cup}{{\raisebox{.5ex} {\smathcalebox{.4}{\ensuremath{\leftarrow}}}}}}
\newcommand{\bigcuparrow}{\charfusion[\mathop]{\bigcup}{\leftarrow}}
\newcommand{\finiteunion}{\cuparrow}
\newcommand{\finitemap}{\ensuremath{\sqsubseteq}}
\newcommand{\dgraph}{dependency graph}
\newcommand{\domain}[1]{{\sc #1}}
\newcommand{\solver}[1]{{\sc #1}}
\providecommand{\problem}[1]{\domain{#1}}
\renewcommand{\v}{\ensuremath{\mathit{v}}}
\providecommand{\vs}[1]{\domain{#1}}
\renewcommand{\vs}{\ensuremath{\mathit{vs}}}
\newcommand{\VS}{\ensuremath{\mathit{VS}}}
\newcommand{\Aut}{\ensuremath{\mathit{Aut}}}
\newcommand{\Inst}[2]{\ensuremath{\mathit{#2 \rightarrow_{#1} #1}}}
\newcommand{\Image}{\ensuremath{\mathit{Im}}}
\newcommand{\Img}[2]{\protect{#1 \llparenthesis #2 \rrparenthesis}}
\newcommand{\SND}{\ensuremath{\mathit{\pi_2}}}
\newcommand{\FST}{\ensuremath{\mathit{\pi_1}}}
\newcommand{\tvsal}{{\pitchfork}}
\newcommand{\nauty}{CGIP}

\newcommand{\pwinter}{\ensuremath{\mathit{\bigcap_{pw}}}}

\newcommand{\dom}{\ensuremath{\mathit{\mathcal{D}}}}
\newcommand{\codom}{\ensuremath{\mathcal{R}}}

\newcommand{\map}{\ensuremath{\mathit{map}}}
\newcommand{\BIJEC}{\ensuremath{\mathit{bij}}}
\newcommand{\INJ}{\ensuremath{\mathit{inj}}}
\newcommand{\funion}{\ensuremath{\overset{\leftarrow}{\cup}}}

\newcommand{\ifnew}{\mbox{\upshape \textsf{if}}}
\newcommand{\thennew}{\mbox{\upshape \textsf{then}}}
\newcommand{\elsenew}{\mbox{\upshape \textsf{else}}}
\newcommand{\choice}{\mbox{\upshape \textsf{ch}}}
\newcommand{\arbchoice}{\mbox{\upshape \textsf{arb}}}
\newcommand{\acycchoice}{\mbox{\upshape \textsf{ac}}}
\newcommand{\cycchoice}{\mbox{\upshape \textsf{cyc}}}
\newcommand{\filter}{\ensuremath{\mathit{FIL}}}
\newcommand{\probset}{\ensuremath{\boldsymbol \Pi}}
\newcommand{\probleq}{\ensuremath{\leq_\Pi}}
\newcommand{\CommVar}{\ensuremath{\bigcap_\v} }
\newcommand{\quotfun}{\ensuremath{ \mathcal{Q}}}

\newcommand{\apre}{\mbox{\upshape \textsf{pre}}}
\newcommand{\aeff}{\mbox{\upshape \textsf{eff}}}
\newcommand{\problist}{\ensuremath \probset}
\newcommand{\cat}{{\frown}}
\newcommand{\probproj}[2]{{#1}{\downharpoonright}^{#2}}
\newcommand{\preced}{\mathbin{\rotatebox[origin=c]{180}{\ensuremath{\rhd}}}}
\newcommand{\perm}{\ensuremath{\sigma}}
\newcommand{\invariant}[2]{\ensuremath{\mathit{inv({#1},{#2})}}}
\newcommand{\invstates}[1]{\ensuremath{\mathit{inv({#1})}}}
\newcommand{\probss}[1]{{\mathcal S}(#1)}
\newcommand{\parChildRel}[3]{\ensuremath{\negdep{#1}{#2}{#3}}}
\newcommand{\asessymbol}{\ensuremath{\mathbb{A}}}
\newcommand{\ases}[1]{{#1}^*}
\newcommand{\uniStates}{\ensuremath{\mathbb{U}}}
\newcommand{\recurrenceDiam}{\ensuremath{\mathit{rd}}}
\newcommand{\recurrenceAcycDiamfun}{\ensuremath{\mathit{{\mathfrak A}}}}
\newcommand{\recurrenceDiamfun}{\ensuremath{\mathit{\mathfrak R}}}
\newcommand{\traversalDiam}{\ensuremath{\mathit{td}}}
\newcommand{\traversalDiamfun}{\ensuremath{\mathit{\mathfrak T}}}
\newcommand{\isPrefix}[2]{\ensuremath{#1 \preceq #2}}
\providecommand{\path}{\ensuremath{\gamma}}
\newcommand{\aspath}{\ensuremath{\vec{\path}}}
\renewcommand{\path}{\ensuremath{\gamma}}
\newcommand{\n}{\textsf{\upshape n}}
\providecommand{\graph}{}
\providecommand{\cal}{}
\renewcommand{\cal}{}
\renewcommand{\graph}{{\cal G}}
\newcommand{\undirgraph}{{\cal G}}
\newcommand{\sset}{\ensuremath{\mbox{\upshape \textsf{ss}}}}
\renewcommand{\ss}{\ensuremath{\state s}}
\newcommand{\slist}{\ensuremath{\vec{\mbox{\upshape \textsf{ss}}}}}
\newcommand{\sll}{\ensuremath{\vec{\state}}}
\newcommand{\listset}{\mbox{\upshape \textsf{set}}}
\newcommand{\listmset}{\mbox{\upshape \textsf{mset}}}
\newcommand{\asset}{\ensuremath{\mathit{K}}}
\newcommand{\aslist}{\ensuremath{\mathit{\overset{\rightarrow}{\gamma}}}}
\newcommand{\head}{\mbox{\upshape \textsf{hd}}}
\renewcommand{\max}{\textsf{\upshape max}}
\newcommand{\argmax}{\textsf{\upshape argmax}}
\renewcommand{\min}{\textsf{\upshape min}}
\newcommand{\bool}{\mbox{\upshape \textsf{bool}}}
\newcommand{\last}{\mbox{\upshape \textsf{last}}}
\newcommand{\front}{\mbox{\upshape \textsf{front}}}
\newcommand{\rot}{\mbox{\upshape \textsf{rot}}}
\newcommand{\stuff}{\mbox{\upshape \textsf{intlv}}}
\newcommand{\tail}{\mbox{\upshape \textsf{tail}}}
\newcommand{\ngrtoas}{\ensuremath{\mathit{\as_{\graph_\mathbb{N}}}}}
\newcommand{\vsfun}{\mbox{\upshape \textsf{vs}}}
\newcommand{\inits}{\mbox{\upshape \textsf{init}}}
\newcommand{\satprecondas}{\mbox{\upshape \textsf{sat-pre}}}
\newcommand{\remcondlessact}{\mbox{\upshape \textsf{rem-cless}}}
\providecommand{\state}{}
\renewcommand{\state}{x}
\newcommand{\statea}{\ensuremath{x_1}}
\newcommand{\stateb}{\ensuremath{x_2}}
\newcommand{\statec}{\ensuremath{x_3}}
\newcommand{\fals}{\mbox{\upshape \textsf{F}}}
\newcommand{\indices}{\ensuremath{V}}
\newcommand{\edges}{\ensuremath{E}}
\newcommand{\vertices}{\ensuremath{V}}
\newcommand{\listtype}{\mbox{\upshape \textsf{list}}}
\newcommand{\settype}{\mbox{\upshape \textsf{set}}}
\newcommand{\pre}{\mbox{\upshape \textsf{pre}}}
\newcommand{\effects}{\mbox{\upshape \textsf{eff}}}
\newcommand{\acttype}{\mbox{\upshape \textsf{action}}}
\newcommand{\graphtype}{\mbox{\upshape \textsf{graph}}}
\newcommand{\projfun}[2]{\ensuremath{\Delta_{#1}^{#2}}}
\newcommand{\snapfun}[2]{\ensuremath{\Sigma_{#1}^{#2}}}
\newcommand{\RDfun}[1]{\ensuremath{{\mathcal R}_{#1}}}
\newcommand{\elldbound}[1]{\ensuremath{{\mathcal LS}_{#1}}}
\newcommand{\distinct}{\textsf{\upshape distinct}}
\newcommand{\ddistinct}{\mbox{\upshape \textsf{ddistinct}}}
\newcommand{\simple}{\mbox{\upshape \textsf{simple}}}

\newcommand{\reachable}[3]{\ensuremath{{#1}\rightsquigarrow{#3}}}

\newcommand{\Omit}[1]{}

\newcommand{\charles}[1]{\textcolor{red}{#1}}
\newcommand{\mohammad}[1]{Mohammad: \textcolor{green}{#1}}

\newcommand{\negreachable}[3]{\ensuremath{{#2}\not\rightsquigarrow{#3}}}
\newcommand{\wdiam}[2]{{#1}^{#2}}
\newcommand{\dsnapshot}[2]{\Delta_{#1}}
\newcommand{\ellsnapshot}[2]{{\mathcal L}_{#1}}

\newcommand{\snapshotsymbol}{|\kern-.7ex\raise.08ex\hbox{\scalebox{0.7}{$\bullet$}}}
\newcommand{\snapshot}[2]{\ensuremath{\mathrel{#1\snapshotsymbol_{#2}}}}
\newcommand{\vstype}{\texttt{\upshape VS}}
\newcommand{\vtype}{{\scriptsize \ensuremath{\dom(\delta)}}}
\newcommand{\Balgo}{{\mbox{\textsc{Hyb}}}}
\newcommand{\ssgraph}[1]{\graph_\ss}
\newcommand{\agree}{\textsf{\upshape agree}}
\newcommand{\ck}{\ensuremath{\texttt{ck}}}
\newcommand{\lk}{\ensuremath{\texttt{lk}}}
\newcommand{\gr}{\ensuremath{\texttt{gr}}}
\newcommand{\gk}{\ensuremath{\texttt{gk}}}
\newcommand{\CK}{\ensuremath{\texttt{CK}}}
\newcommand{\LK}{\ensuremath{\texttt{LK}}}
\newcommand{\GR}{\ensuremath{\texttt{GR}}}
\newcommand{\GK}{\ensuremath{\texttt{GK}}}
\newcommand{\safe}{\ensuremath{\texttt{s}}}

\newcommand{\derivname}{\ensuremath{\partial}}
\newcommand{\deriv}[3]{\ensuremath{\derivname(#1,#2,#3)}}
\newcommand{\derivabbrev}[3]{\ensuremath{{\partial(#1,#2)}}}
\newcommand{\subsetoracle}{\ensuremath{ \Omega}}
\newcommand{\Aalgo}{{\mbox{\textsc{Pur}}}}
\newcommand{\Sname}{\textsf{\upshape S}}
\newcommand{\Sbrace}[1]{\Sname\langle#1\rangle}
\newcommand{\SalgoName}{\Sname_{\textsf{\upshape max}}}
\newcommand{\Salgo}[1]{\SalgoName\langle#1\rangle}

\newcommand{\WLPname}{{\mbox{\textsc{wlp}}}}
\newcommand{\WLPbrace}[1]{\WLPname\langle#1\rangle}
\newcommand{\WLPalgoName}{\WLPname_{\textsf{\upshape max}}}
\newcommand{\WLP}[1]{\WLPalgoName\langle#1\rangle}

\newcommand{\Nname}{\ensuremath{\textsf{\upshape N}}}
\newcommand{\Nbrace}[1]{\Nname\langle#1\rangle}
\newcommand{\NalgoName}{\Nname{_{\textsf{\upshape sum}}}}
\newcommand{\Nalgobrace}[1]{\NalgoName\langle#1\rangle}

\newcommand{\acycNname}{\widehat{\textsf{\upshape N}}}
\newcommand{\acycNbrace}[1]{\acycNname\langle#1\rangle}
\newcommand{\acycNalgoName}{\acycNname{_{\textsf{\upshape sum}}}}
\newcommand{\acycNalgobrace}[1]{\acycNalgoName\langle#1\rangle}

\newcommand{\Mname}{\ensuremath{\textsf{\upshape M}}}
\newcommand{\Mbrace}[1]{\Mname\langle#1\rangle}
\newcommand{\MalgoName}{\Mname{_{\textsf{\upshape sum}}}}
\newcommand{\Malgobrace}[1]{\MalgoName\langle#1\rangle}
\newcommand{\cardinality}[1]{{\ensuremath{|#1|}}}
\newcommand{\length}[1]{\cardinality{#1}}
\newcommand{\basecasefun}{\ensuremath{b}}
\newcommand{\Basecasefun}{\ensuremath{\mathcal B}}

\newcommand{\vertexgen}{\ensuremath{u}}
\newcommand{\vertexa}{{\ensuremath{\vertexgen_1}}}
\newcommand{\vertexb}{{\ensuremath{\vertexgen_2}}}
\newcommand{\vertexc}{{\ensuremath{\vertexgen_3}}}
\newcommand{\vertexd}{{\ensuremath{\vertexgen_4}}}
\newcommand{\vertexsetgen}{\ensuremath{\mathit{us}}}
\newcommand{\vertexseta}{\vertexsetgen_1}
\newcommand{\vertexsetb}{\vertexsetgen_2}
\newcommand{\labelsymbol}{\ensuremath{l}}
\newcommand{\labelfun}{\ensuremath{\mathcal{L}}}
\newcommand{\DAG}{\ensuremath{A}}
\newcommand{\NalgoNameN}{{\ensuremath{\NalgoName_{\mathbb{N}}}}}
\newcommand{\NnameN}{\ensuremath{\Nname_\mathbb{N}}}
\newcommand{\replaceprojsinglename}{\raisebox{-0.3mm} {\scalebox{0.7}{\textpmhg{H}}}}
\newcommand{\replaceprojsingle}[3] {{ #2} \underset {#1} {\raisebox{-0.3mm} {\scalebox{0.7}{\textpmhg{H}}}}  #3}
\newcommand{\HOLreplaceprojsingle}[1]{\underset {#1} {\raisebox{-0.3mm} {\scalebox{0.7}{\textpmhg{H}}}}}

\newcommand{\lotus}{{\scalebox{0.6}{\includegraphics{lotus.pdf}}}}
\newcommand{\invlotus}{\mathbin{\rotatebox[origin=c]{180}{$\lotus$}}}
\newcommand{\clique}{\ensuremath{K}}
\newcommand{\partition}{\ensuremath{\vs_{1..n}}}
\newcommand{\partitiontype}{\ensuremath{\vstype_{1..n}}}
\newcommand{\vtxpartition}{\ensuremath{P}}

\newcommand{\traversalDiamAlgo}{{\mbox{\textsc{TravDiam}}}}
\newcommand{\prefix}{\textsf{\upshape pfx}}
\newcommand{\powerset}{\mathbb{P}}
\newcommand{\postfix}{\textsf{\upshape sfx}}
\newcommand{\dfunproj}{\ensuremath{{\mathfrak D}}}
\newcommand{\dfunsnap}{\ensuremath{{\textgoth D}}}
\newcommand{\ellfunproj}{\ensuremath{\mathfrak L}}
\newcommand{\ellfunsnap}{\ensuremath{\textgoth L}}
\newcommand{\cycle}{\ensuremath{C}}
\newcommand{\petal}{\ensuremath{\eta}}
\renewcommand{\prod}{\ensuremath{{{{{\mathlarger{\mathlarger {{\mathlarger {\Pi}}}}}}}}}}
\newcommand{\sccset}{{\ensuremath{SCC}}}
\newcommand{\scc}{{\ensuremath{scc}}}
\newcommand{\negate}[1]{\overline{#1}}
\newcommand{\setofsets}{\ensuremath{S}}
\newcommand{\group}{\ensuremath{\cal \Gamma}}
\newcommand{\neededvars}{{\cal N}}
\newcommand{\sspace}{\mbox{\upshape \textsf{sspc}}}
\newcommand{\tip}{\ensuremath{t}}
\newcommand{\vara}{\ensuremath{\v_1}}
\newcommand{\varb}{\ensuremath{\v_2}}
\newcommand{\varc}{\ensuremath{\v_3}}
\newcommand{\vard}{\ensuremath{\v_4}}
\newcommand{\vare}{\ensuremath{\v_5}}
\newcommand{\varf}{\ensuremath{\v_6}}
\newcommand{\varg}{\ensuremath{\v_7}}
\newcommand{\varh}{\ensuremath{\v_8}}
\newcommand{\vari}{\ensuremath{\v_9}}
\newcommand{\acta}{\ensuremath{\act_1}}
\newcommand{\actb}{\ensuremath{\act_2}}
\newcommand{\actc}{\ensuremath{\act_3}}
\newcommand{\actd}{\ensuremath{\act_4}}
\newcommand{\acte}{\ensuremath{\act_5}}
\newcommand{\actf}{\ensuremath{\act_6}}
\newcommand{\actg}{\ensuremath{\act_7}}
\newcommand{\acth}{\ensuremath{\act_8}}
\newcommand{\acti}{\ensuremath{\act_9}}

\newcommand{\planningproblem}{\Uppi}

\tikzset{dots/.style args={#1per #2}{line cap=round,dash pattern=on 0 off #2/#1}}
\providecommand{\moham}[1]{\fbox{{\bf \@Mohammad: }#1}}
\newcommand{\TDbound}{{\mbox{\textsc{Arb}}}}
\newcommand{\expbound}{{\mbox{\textsc{Exp}}}}
\newcommand{\sasdom}{\expbound}
\newcommand{\cardfun}{\ensuremath{\mathbb{C}}}
\newcommand{\AGNa}{AGN1}
\newcommand{\AGNb}{AGN2}
\newcommand{\reset}{{\ensuremath{reset}}}
\newcommand{\cost}{\mathcal{C}}
\newcommand{\goal}{\mathcal{G}}
\newcommand{\init}{\mathcal{I}}
\newcommand{\completenessthreshold}{\textit{CT}}
\newcommand{\subsetDiam}{\mathcal{S}}
\newcommand{\automorphism}{\mathcal{A}}
\newcommand{\orbits}{\mathcal{O}}
 \renewcommand{\mohammad}[1]{}
\renewcommand{\fred}[1]{}

\renewcommand{\as}{\textit{ops}}
\renewcommand{\childrensymbol}{\textsf{\upshape child}}
\renewcommand{\traversalDiamAlgo}{{\mbox{\textsc{TravD}}}}
\renewenvironment{isabelle}{
}{
    \ignorespacesafterend
}
\renewenvironment{isabellebody}{
}{
    \ignorespacesafterend
}
\renewenvironment{isamarkuptext}{
}{
    \ignorespacesafterend
}

\newcommand{\isabellefun}[1]{{\small \texttt{#1}}}
\newcommand{\citeyearparen}[1]{(\citeyear{#1})}
\renewcommand{\equiv}{=}
\section{Introduction}
\label{sec:intro}

Planning systems are becoming more and more scalable and efficient, as shown by different planning competitions~\cite{ipc98,DBLP:journals/aim/ColesCOJLSY12,vallati20152014}, making them suited for realistic applications.
Since many applications of planning are safety-critical, increasing the trustworthiness of planning algorithms and systems is instrumental to their widespread adoption.
Consequently there currently are substantial efforts to improve the trustworthiness of planning systems~\cite{howey2004val,eriksson2017unsolvability,abdulaziz2018formally,ictai2018}.

Increasing trustworthiness of software is a well-studied problem.
Three approaches have been tried in the literature~\cite{DBLP:conf/mfcs/AbdulazizMN19}.
Firstly, a system's trustworthiness can be increased by applying software engineering techniques, e.g.\ programming at the right level of abstraction, code reviewing, and testing.
Although these practices are relatively easy to implement, they are incomplete.

Secondly, there is certified computation, where the given program computes, along with its output, a certificate showing why this output is correct. 
This relegates the burden of trustworthiness to the certificate checker, which should be much simpler than the system whose output is to be certified, and thus is less error prone.
Certified computation was pioneered by~\citeauthor{mehlhorn98mfcs}~\citeyear{mehlhorn98mfcs} who used it for their LEDA library.
In the realm of planning, this approach was pioneered by~\citeauthor{howey2004val} who developed the plan validator VAL~\cite{howey2004val}.
Also, certifying unsolvability for planning was pioneered by~\citeauthor{eriksson2017unsolvability}~\citeyear{eriksson2017unsolvability}.
A fundamental problem with certified computation for classical planning, as well as other PSPACE-complete problems, is that no succinct certificates exist unless $\mathrm{NP} = \mathrm{PSPACE}$~\cite[Chapter 4]{DBLP:books/daglib/0023084}.

Thirdly, there is formal verification, where system properties are verified by means of a mechanically checkable formal mathematical proof.
This gives the highest possible trust in a program.
Unlike certified computation, formal verification guarantees program completeness, in addition to its output correctness.
Nonetheless, formal verification needs intensive effort, and it is usually much harder for a formally verified program to perform as efficiently as an unverified program since verifying all performance optimisations is usually infeasible.
Nonetheless, formal verification has seen recent wide-spread success.
Notable applications include a verified OS kernel~\cite{klein2009sel4}, a verified SAT-solver~\cite{DBLP:journals/jar/BlanchetteFLW18}, a verified model checker~\cite{esparza2013fully}, a verified conference system~\cite{DBLP:conf/cav/KanavL014}, a verified optimised C compiler~\cite{leroy2009formal}, and, in the context of planning, verified validators~\cite{ictai2018,TemporalSemantics}, and algorithms for bounding plan lengths~\cite{abdulaziz2018formally}.

Encoding planning problems as logical formulae has a long history~\cite{mccarthy1981some}.
SAT-based planning~\cite{kautz:selman:92} is the most successful such approach, where the question of whether a planning problem has a solution of bounded length is encoded into propositional (SAT) formulae.
For a sound and complete encoding, a formula for a length bound (aka horizon) $h$ is satisfiable iff there is a solution with length bounded by $h$.
A series of such formulae for increasing horizons are passed to a SAT solver until a plan is found.
The encoding of~\citeauthor{kautz:selman:92} was significantly improved by exploiting problem symmetries~\cite{rintanen2003symmetry}, state invariants, and by parallelising the encodings~\cite{rintanen:06}.
In this paper we present a formally verified $\forall$-step parallel encoding of classical planning~\cite{kautz:selman:96,rintanen:06}.
Using it as a reference, we discover bugs in the state-of-the-art SAT-based planner Madagascar~\cite{rintanen:12}.
We also show that our encoder is an order of magnitude slower than Madagascar, which is reasonable for formally verified software.

\paragraph{Motivation} Before we delve into details, we find it sensible to answer the question of why to verify a SAT encoding of planning?
The first motivation is to demonstrate the process of formally verifying a state-of-the-art planning algorithm.
Secondly, a verified planner can be used as a reference implementation, against which unverified planners can be tested.
Thirdly, SAT encodings of planning are particularly suited for formal verification since the process of encoding the problem into a SAT formula is not the computational bottleneck.
This means that not much effort needs to be put into verifying different implementation optimisations.
Also, importantly, verified SAT solving technology is continuously advancing, which means that the performance of verified planning using the verified encoding will continuously improve.
E.g.\ the verified SAT solver by~\citeauthor{DBLP:journals/jar/BlanchetteFLW18} is one to two orders of magnitude slower than Minisat~\cite{DBLP:conf/sat/EenS03}, and the verified unsatisfiability certificate checker GRAT~\cite{DBLP:conf/sat/Lammich17} is faster than DRAT-trim~\cite{DBLP:conf/sat/WetzlerHH14}.
Lastly, the problem of increasing trustworthiness of `traditional' SAT-based planning was most recently raised by~\citeauthor{DBLP:conf/aips/ErikssonH20}~(\citeyear{DBLP:conf/aips/ErikssonH20}).
They stated, as an open problem, certifying that an unsatisfiable SAT formula encoding a planning problem indeed shows a lower bound on the solution length.
We address this open problem and show it is solvable with formal verification.

\paragraph{Approach Overview} Our approach has two focuses, which we try to highlight in this paper.
The first is \emph{minimising trusted code}.
To reduce the trusted code base we make sure the encoder and the decoder take (produce) as input (output) abstract syntax trees (AST) of the formats used in the input (output) files.
This way we only trust parsing and pretty printing components.
The encoder takes an AST of Fast-Downward's translator format~\cite{Helmert06} as input (hereafter, FD-AST) and produces an AST of the standard DIMACS-CNF format.
This DIMACS-CNF is passed to an unverified SAT solver.
If the SAT solver finds a model for the encoded formula, our verified decoder takes the AST of the DIMACS model and the FD-AST and produces a plan only if the model entails the CNF encoding of the FD-AST.
If the SAT solver finds the formula unsatisfiable, we take the unsatisfiability certificate and pass it, together with the DIMACS-CNF encoding, to the formally verified unsatisfiability certificate checker of~\citeauthor{DBLP:conf/sat/Lammich17}~\citeyear{DBLP:conf/sat/Lammich17}.
Accordingly, all outputs of our system have formal guarantees, whether the output is a plan or a conclusion that none exists.

The second is \emph{engineering tradeoffs for feasible verification}.
Formally verifying a reasonably big piece of software is a daunting task.
E.g.\ it has been reported that the size of the proof scripts for verifying the OS kernel \emph{seL4} is quadratically as big as the C implementation of the kernel~\cite{klein2009sel4}.
Thus, a major verification effort like verifying a SAT-based planner needs careful engineering and modularisation.
For complicated algorithms, the most successful approach is stepwise refinement, where one starts with an abstract version of the algorithm and verifies it.
Then one devises more optimised versions of the algorithm, and only proves the optimisations correct.
This approach was used in most successful algorithm verification efforts~\cite{klein2009sel4,DBLP:journals/jar/BlanchetteFLW18,esparza2013fully,DBLP:conf/cav/KanavL014}.

For compilers, a different approach is used: one splits the compiler into smaller translation steps, and verifies each one of those steps separately.
In the end, the composition of those verified transformations is the verified compiler.
This approach was used in all notable verified compilers~\cite{leroy2009formal,DBLP:conf/popl/KumarMNO14}.
We follow this methodology. 
\spacex{In particular, for the encoding direction we prove translation steps from FD-AST into a more restricted finite domain representation (FDR) of planning problems, then to STRIPS, next to propositional logic formulas and lastly to DIMACS-AST.
Decoding starts with a model of the DIMACS encoding, which is decoded to a STRIPS parallel solution then to a FDR serial solution and lastly to a FD-AST serial solution.}{}

\paragraph{Isabelle/HOL} We perform the verification using the interactive theorem prover Isabelle/HOL~\cite{DBLP:books/sp/NipkowPW02}, which is a theorem prover for Higher-Order Logic. 
Roughly speaking, Higher-Order Logic can be seen as a combination of functional programming with logic.
Isabelle is designed for trustworthiness: following the Logic for Computable Functions approach (LCF)~\cite{milner1972logic}, a small kernel implements the inference rules of the logic.
Around the kernel there is a large set of tools that implement proof tactics and high-level concepts like algebraic datatypes and recursive functions.
Bugs in these tools cannot lead to inconsistent theorems being proved, but only to error messages when the kernel refuses a proof.

\paragraph{Availability}
All theorems in this paper were formally proved in Isabelle/HOL.
\fullversion{We attach an appendix describing the formalisation of interesting definitions and concepts.}
The full formal proofs can be found at \url{https://www.isa-afp.org/entries/Verified_SAT_Based_AI_Planning.html} and the verified planner based on these proofs can be found at \url{https://github.com/mabdula/Verified-SAT-Based-Planning}.
 
\begin{figure}
\begin{minipage}{.13\textwidth}
\centering
{\tiny
\begin{verbatim}
1
begin_variable
var1
-1
2
Atom at-robby(r0)
Atom at-robby(r1)
end_variable
\end{verbatim}
}
\end{minipage}
\begin{minipage}{.1\textwidth}
\centering
{\tiny
\begin{verbatim}
begin_state
0
end_state
begin_goal
1
0 1
end_goal

\end{verbatim}
}
\end{minipage}
\begin{minipage}{.12\textwidth}
\centering
{\tiny
\begin{verbatim}
1
begin_operator
move r0 r1
0
1
0 0 0 1
0
end_operator
\end{verbatim}
}
\end{minipage}
\begin{minipage}{.1\textwidth}
\centering
{\tiny
\begin{linenumbers}
\begin{verbatim}
p cnf 6 15
3 0
-5 0
-4 0
...
5 -6 1 0
-3 4 1 0
-5 6 0
-1 1 0
\end{verbatim}
\end{linenumbers}}
\end{minipage}
\caption{The first three listings are the concrete syntax of a planning problem in Fast Downward's translator format.
The fourth listing is a CNF-DIMACS formula.
\vspace{-3ex}
\label{fig:concretesyntax}}
\end{figure}
\section{Background}
\label{sec:bg}

In this paper, lists (sets) of objects are written between square brackets (curly braces).
E.g.\  $[a,b,c]$ ($\{a,b,c\}$) denotes the list (set) of objects $a$, $b$ and $c$.
We also make use of the choice function $\choice$, which, for a non-empty set $s$, denotes an arbitrary element of $s$.
If $s = \{\}$, $\choice$ is undefined.
A mapping $f:V\rightarrow A$ is a set of maplets, s.t., for every $v\mapsto a\in f$,  $v \in V$ and $a\in A$, and, if $v\mapsto a_1\in f$ and $v\mapsto a_2\in f$, then $a_1 = a_2$.
For a mapping $f$, we define $f(v)$ to be $a$ if $v\mapsto a\in f$, otherwise it is undefined.
Also, for a mapping $f:V\rightarrow A$, $\dom(f)$ denotes $\{v \mid v\mapsto a \in f\}$.
We call $f$ a complete mapping iff $\dom(f) = V$.
Otherwise, we call it a partial mapping.

There are multiple formalisms to represent planning and logical formulae.
Some of those formalisms are geared towards being file formats and others are abstract formalisms that are geared towards performing abstract and pen-and-paper reasoning.
Since the main goal of our development is to minimise trusted (i.e.\ unverified) code, we aim at having our verified program \begin{enumerate*} \item take as input as close a representation as possible to the actual input format (i.e.\ Fast-Downward's translator format\footnote{http://www.fast-downward.org/TranslatorOutputFormat}) and \item produce as output a format that is as close to the output file format as possible (i.e.\ the DIMACS-CNF format).\end{enumerate*}
However, although we want our verified program to operate on inputs (outputs) that are as close as possible to the input (output) files, it cannot directly operate on the input (output) strings.
In particular, the input strings have to be parsed into ASTs, and the program has to produce an output that is also an AST, which is to be pretty printed into an output file.

\newcommand{\varsec}{\ensuremath{V}}
\newcommand{\opname}{\ensuremath{\textit{name}}}
\newcommand{\opaxiom}{\ensuremath{\textit{prev}}}
\newcommand{\oppres}{\ensuremath{\textit{ps}}}
\newcommand{\opeffs}{\ensuremath{\textit{es}}}
\newcommand{\effpre}{\ensuremath{\textit{epre}}}

\begin{mydef}[FD-AST]
\label{def:FDAST}
An FD-AST is a tuple $\langle V,I,G,O\rangle$, where $V:\mathbb{N}\rightarrow \mathbb{N}$ is the variables section, $I:\mathbb{N}\rightarrow \mathbb{N}$ is the initial state, $G:\mathbb{N}\rightarrow \mathbb{N}$ is the goal, and $O$ is the set of operators.
For each $v \mapsto D_v \in V$, $v$ stands for a variable and $D_v$ is the number of assignments this variables can take.
An operator is a tuple $\langle \opname, \oppres, \opeffs\rangle$, where $\opname$ is the operator's name, $\oppres:\mathbb{N}\times\mathbb{N}$ is a partial mapping called the prevail preconditions, and $\opeffs$ is a set of effects.
An effect is a tuple $\langle \effpre, v, l, m\rangle$, where $\effpre:\mathbb{N}\rightarrow\mathbb{N}$ is a partial mapping called the effect's precondition and $v,l,m\in\mathbb{N}$.
If $\effpre$ is non-empty the effect is called conditional.
\end{mydef}

\noindent Note: in parsing, we ignore parts of the format that are irrelevant to our encoding, e.g.\ metric, mutex, and axiom sections.

\begin{mydef}[Well-Formedness]
\label{def:FDASTwf}
An FD-AST $\langle V, I, G, O \rangle$ is well-formed iff \begin{enumerate*} \item $I$ is a well-formed state w.r.t.\ the problem, \item $G$ is a well-formed partial state w.r.t.\ the problem, \item all operators have distinct names, and \item every operator is well-formed  w.r.t.\ the problem.\end{enumerate*}
A (partial) state $s$ is well-formed w.r.t.\ a problem $\langle V, I, G, O \rangle$ iff it is a (partial) mapping s.t.\ $\dom(s)\subseteq\dom(\varsec)$, and for any $v\in\dom(s)$, $s(v) < \varsec(v)$.
An operator $\langle \opname, \oppres, \opeffs\rangle$ is well-formed iff \begin{enumerate*} \item $\oppres$ is a well-formed partial state, \item $v_1\neq v_2$, for every $\langle \effpre_1, v_1, l_1, m_1\rangle, \langle \effpre_2, v_2, l_2, m_2\rangle\in \opeffs$ and \item for every $\langle \effpre, v, l, m\rangle \in \opeffs$: \begin{enumerate*}[label*=(\arabic*)] \item $\effpre$ is a well-formed partial state, and \item $l,m < \varsec(v)$.\end{enumerate*} \end{enumerate*}
\end{mydef}

\begin{mydef}[Execution]
\label{def:FDASTex}
An operator $\langle \opname, \oppres, \opeffs\rangle$ is executable in a state $s$ iff $\oppres \subseteq s$ and, for each effect $\langle \effpre, v, l, m\rangle\in \opeffs$, $v\mapsto l\in s$, if $0 \leq l$.
The state resulting from executing the operator is 
{\renewcommand{\theequation}{\roman{equation}}
\setcounter{equation}{0}
 \begin{multline*}
(s\setminus \{v \mapsto l\mid \exists\langle\effpre, v, l, m\rangle\in \opeffs \wedge \effpre \subseteq s\}) \cup\\
 \{v \mapsto m\mid \exists \langle\effpre, v, l, m\rangle\in \opeffs \wedge \effpre \subseteq s\},
\end{multline*}}
\noindent i.e.\ it is the same as $s$, except that it assigns variables according to the effects whose preconditions are satisfied.

A solution to a problem $P$ is a list of names $[\opname_1, \opname_2, \dots \opname_n]$ s.t.\ there is a mapping $\textit{sol}:\{\opname_1\dots \opname_n\}\rightarrow O$ s.t.\ $\textit{sol}(\opname_i)\in O$, for all $1\leq i\leq n$, and the sequence of operators $[\textit{sol}(\opname_1), \textit{sol}(\opname_2), \dots \textit{sol}(\opname_n)]$ are executable in-order starting at $I$, and the goal is a subset of the state resulting from executing all operators.
\end{mydef}

\begin{myeg}
\label{eg:FDAST}
Figure~\ref{fig:concretesyntax} shows a planning problem in Fast Downward's translator syntax that models a robot that is in one room and whose goal is to move to another room. 
The abstract syntax tree of that problem is $\langle[\langle v_0,2\rangle],\{v_0\mapsto 0\}, \{v_0\mapsto 1\}, \langle \textit{move}, 0, \{\}, \{\langle[], v_0, 0, 1\rangle\}\rangle\rangle$.
A solution to this problem is $[\textit{move}]$.
\end{myeg}
\noindent We limit ourselves to problems without conditional effects and with only consistent preconditions, defined as follows.
\begin{mydef}[Valid FD-AST Problem]
\label{def:validP}
We call a problem $\langle V, I, G, O \rangle$ to be valid iff \begin{enumerate*} \item it is well-formed, \item $\effpre=\emptyset$, for any $\langle \opname, \oppres, \opeffs\rangle\in O$ and $\langle \effpre, v, l, m\rangle \in \opeffs$, i.e.\ it has no conditional effects, and \item for any variable $v\in\dom(p_1)\cap\dom(p_2)$ where $p_1,p_2 \in \{\oppres\}$, where $\langle \opname, \oppres, \opeffs\rangle\in O$, we have $p_1(v) = p_2(v)$, i.e.\ all preconditions are consistent.\end{enumerate*}
\end{mydef}

To formally verify our encoding, we need to formalise the definitions in Isabelle/HOL's logic.
As input, we use the formalisation of FD-AST developed by \citeauthor{ictai2018}~\citeyear{ictai2018}.
\fullversion{Listing~\ref{isa:FDAST} in the appendix and the associated description review the formalisation of FD-AST.}

Note that this representation of planning problems, as well as the final SAT formula, have features which make it is too cumbersome for abstract pen-and-paper reasoning.
However, we need to start from representations which are as close as possible to the input/output file to reduce trusted code.
In particular, if we prove our encoding to be correct on other more abstract representations, we will have to use trusted (i.e.\ unverified) pre-processing programs to convert between the files and the more abstract representation.
We also note that this software is only a simple parser if one considers the input to be in FD's translator's format, which is our claimed input.
The trusted code base is significantly larger if one considers the input to be a PDDL domain and instance, as the conversion then includes grounding and the computation of invariants, etc.

The rest of the paper is structured s.t.\ there is a section describing every intermediate representation and translation step, with the associated correctness theorem.
 \section{Translating FD-AST to FDR}
\label{sec:FDtoSASP}

Although we can define our encoding directly on FD-AST, we opted to firstly translate the FD-AST to the Finite Domain Representation (FDR), which is another representation of planning problems with multi-valued state variables.
FDR is more abstract than FD-AST.
This facilitates smoother formal reasoning, e.g.\ it is more suitable for stating algorithms and theorem statements, while FD-AST is a file format.
In most expositions in planning literature, there is not a distinction between file formats and the abstract formalisms on which algorithms and theorems are stated.
The actual implementations, however, start from a file format, like Fast Downward's translator format, which is simplified to a more abstract formalism.
In our case, nonetheless, we make the distinction between the two formalisms since our main goal is a formal correctness guarantee on an implementation of a planner, including the translation of FD-AST to FDR.
\newcommand{\range}{\ensuremath{\mathcal{R}}}
\newcommand{\execact}{\ensuremath{\gg_+}}
\renewcommand{\pre}{\textit{p}}
\newcommand{\eff}{\textit{e}}
\newcommand{\fdrop}{\textit{op}}
\newcommand{\fdrops}{\textit{ops}}
\fred{Parallel execution semantics pen \& paper and Isabelle formalisation are missing}
\renewcommand{\state}{s}
\begin{mydef}[Finite Domain Representation]
\label{def:FDR}
An FDR planning problem $\Psi$ is a tuple $\langle V,\mathcal{R},I,O,G\rangle$, where $V$ is the set of state variables, $\range:V\rightarrow A_v$ is mapping from variables to sets of assignments, $O$ is a set of operators, $I$ is the initial state, and $G$ is the goal.
Each variable (operator) has a unique natural number index, where, for a variable $v\in V$ (operator $\fdrop\in O$), its index is $0\leq v_i < \cardinality{V}$ ($0\leq \fdrop_i < \cardinality{O}$).
For FDR, a (partial) state $\state:V\rightarrow \bigcup \{\mathcal{R}(v)\mid v\in V\}$ is a mapping, and $\dom(\state)\equiv \{\v\mid \v\mapsto a\in\state\}$.
A (partial) state $\state$ is valid iff for any $\v\mapsto a\in\state$ we have that $a\in\range(v)$.
A valid operator is a pair of valid partial states $\fdrop\equiv\langle p,e\rangle$, where $p$ is the precondition, denoted by $\pre(\fdrop)$, and $e$ is the effect, denoted by $\eff(\fdrop)$.
\fred{Is the definition for $\state\execact \fdrop$ really adequate? E.g. what if $e=\{\}$?}
We denote the the execution of a sequence of operators $\fdrops$ at a state $\state$ by $\exec{\state}{\fdrops}$, and it is defined as follows: if $\fdrops$ is not empty and if for the first operator $\langle p,e\rangle$ in $\fdrops$ we have $p\subseteq\state$ then $\exec{\state}{\fdrops}\equiv \exec{\exec{\state}{\fdrop}}{\fdrops'}$, where $\fdrops'$ is the tail (i.e.\ every element but the first) of $\fdrops$ and $\exec{\state}{\fdrop}$ is defined to be $\{\v\mapsto a\mid \ifnew\; \v\in\dom(e)\;\thennew\;a=e(\v)\;\elsenew\;\state(\v)\}$.
Otherwise, $\exec{\state}{\fdrops}\equiv\state$.
$\Psi$ is a valid FDR problem iff $I$ is a valid state, $G$ is a valid partial state, and $O$ is a set of valid operators.
A solution for $\Psi$ is an operator sequence $\fdrops$ where all operator in $\fdrops$ come from $O$ and $G\subseteq \exec{I}{\fdrops}$.
\end{mydef}
\noindent Note: in contrast to FD-AST plan execution semantics, FDR execution semantics are defined as a total function, where plan execution always returns the last state reached before the first operator whose preconditions are not satisfied.
A total execution function makes many of the formal proofs easier.

Translating FD-ASTs into FDR problems is done using the following encoding.
\newcommand{\asttofdr}{\ensuremath{\textit{FDR}}}
\newcommand{\fdrtoast}{\ensuremath{AST}}
\begin{myenc}
\label{enc:FD-ASTtoFDR}
For an FD-AST problem $P\equiv\langle V,I,G,O\rangle$, let $\asttofdr_V\equiv\{v\mid\exists n.\;\langle v,n\rangle\in V\}$ and $\asttofdr_\mathcal{R}\equiv\{v\mapsto \{0\dots n-1\}\mid\langle v,n\rangle\in V\}$.
For an FD-AST effect $e\equiv\langle \effpre, v, l, m\rangle$, let $e_{old}$ denote $v\mapsto l$ and $e_{new}$ denote $v\mapsto m$.
For an operator $\fdrop\equiv\langle \opname,\oppres,\opeffs\rangle$, let $\asttofdr_O(\fdrop)\equiv \langle\oppres\cup\{e_\textit{old}\mid e\in \opeffs\}, \{e_{new} \mid e \in \opeffs\}\rangle$.
For the FD-AST problem $P$, its encoding as an FDR problem, $\asttofdr(P)$, is $\langle\asttofdr_V, \asttofdr_\mathcal{R}, I, \{\asttofdr_O(\fdrop)\mid \fdrop\in O\}, G\rangle$.
\end{myenc}
For the other direction of this encoding, we devise a decoding function $\fdrtoast$ that, given a solution for the FDR problem $\asttofdr(P)$, decodes it into a solution for $P$.
\begin{mydec}
\label{dec:FDRtoFD-AST}
First, for a set $s$ let $\choice(s)$ denote an arbitrary element of $s$ if $s$ is not empty, and undefined otherwise.
For an FDR operator $\fdrop$, let $\fdrtoast(\fdrop)\equiv\choice\{\opname\mid \asttofdr_O(\langle \opname,\oppres,\opeffs\rangle) = \fdrop\}$, where $O$ are the operators in $P$.
\end{mydec}

\begin{myeg}
\label{eg:FDR}
The compiled FDR equivalent to the FD-AST problem $P$ in Example~\ref{eg:FDAST}, $\asttofdr(P)$, is 
{\renewcommand{\theequation}{\roman{equation}}
\setcounter{equation}{0}
 \[\left\langle
\begin{gathered}
  \asttofdr_V=\{\v_0\},\asttofdr_\mathcal{R}=\{\v_0 \mapsto\{0,1\}\},\\
   I=\{\v_0\mapsto 0\},\asttofdr_O=\{\fdrop_0\},\asttofdr_G=\{\v_0\mapsto 1\}
\end{gathered}
\right\rangle,\]}
where $\fdrop_0\equiv\langle\{\v_0\mapsto 0\}, \{\v_0 \mapsto 1\}\rangle$.
\end{myeg}

The following theorem represents the soundness and completeness of this compilation step.
\begin{mythm}
\label{thm:asttofdr}
Let $P$ be a valid FD-AST problem.
We have that: \begin{enumerate*} \item if $[\opname_1,\opname_2,\dots]$ is a plan for $P$ then $[\asttofdr_O(\choice(O_{\opname_1}),$ $\asttofdr_O(O_{\opname_2}),\dots]$ is a plan for the FDR task $\asttofdr(P)$, where, for $\opname_i$, $O_{\opname_i}\equiv \{\fdrop \mid \exists\;\oppres'\;\opeffs'. \fdrop \in O \wedge \fdrop = \langle \opname_i,\oppres',\opeffs'\rangle\}$, and \item if $[\fdrop_1,\fdrop_2,\dots]$ is a plan for the FDR task $\asttofdr(P)$, then $[\fdrtoast(\fdrop_1'),\fdrtoast(\fdrop_2'),\dots]$ is a plan for $P$, where $[\fdrop_1',\fdrop_2',\dots]$ are the operators from the given FDR plan whose preconditions are satisfied.
\end{enumerate*} 
\end{mythm}
\begin{proof}[Proof sketch]
Both statements are proved by induction on the length of the given plan, while generalising over the initial state of $P$, and then careful unfolding of Definitions~\ref{def:FDAST}, \ref{def:FDASTwf}, \ref{def:FDASTex}, \ref{def:validP}, \ref{def:FDR}, Encoding~\ref{enc:FD-ASTtoFDR} and Decoding~\ref{dec:FDRtoFD-AST}.
\end{proof}
\noindent Note: since FDR execution function is total, while that of FD-AST is not, operators whose preconditions are not satisfied have to be removed when decoding the FDR plan.

Before we close this section we note a few points regarding the formal proof of the above theorem.
The proof of this theorem does not have complicated mathematical ideas or constructions.
However, the main difficulty is correctly formulating the definitions of well-formed and valid FD-ASTs, valid FDRs and the encoding and the decoding.
Due to the many conjuncts and components of these definitions, their interactions make formally stating these definitions and proving the theorem a very error-prone and cumbersome process.
E.g.\ a detail in the formal proof, which would be glossed over in a pen-and-paper treatment, is to show that encoding a well-formed valid FD-AST results in a valid FDR, which is necessary for using the different theorems about FDR problems.
Proving that depends on the assumption that the AST operators have no conditional effects, a fact which we only understood during our development of the formal proof.
To overcome these difficulties, we employed deliberate engineering efforts to make the formal proof more modular.
E.g.\ we split the encoding of FD-AST operator to FDR operators into multiple stages.
First, the effect preconditions are removed and added into the operator's preconditions, resulting in a simpler FD-AST, with no effect preconditions.
Then, we define a function to encode these simpler FD-ASTs into FDRs.
\fullversion{More details on these proof engineering efforts are in the appendix.}

\section{Translating FDR to STRIPS}

\label{sec:sasptostrips}
\mohammad{Merge all the isabelle formalisation paragraphs in one subesction, and make them a bit shorter.}
\mohammad{Formlulae should be inline.}

Since a SAT encoding only has propositional variables, we need to compile the multi-valued state variables of FDR to propositional values.
Instead of performing the compilation of the multi-valued variables together with the compilation of the transition relation in one step, we opted to do them in separate steps, where we first compile FDR problems to STRIPS problems and then compile STRIPS problems into SAT formulae.
This decision is not of much theoretical importance, but more geared towards making the verification more modular and thus more manageable.
\begin{mydef}[STRIPS Problem]
\label{def:strips}
An FDR problem $\Pi\equiv\langle V,\mathcal{R},I,O,G\rangle$ is a STRIPS problem iff $\mathcal{R}(v)=\{\bot,\top\}$, for all $v\in V$.
When constructing STRIPS mappings, we denote $v \mapsto \top$ with $v$ and $v \mapsto \bot$ with $\negate{v}$.
\end{mydef}

\newcommand{\fdrtostrips}{\ensuremath{\varphi}}
\newcommand{\stripstofdr}{\ensuremath{\varphi^{-1}}}

\begin{myenc}
\label{enc:fdrtostrips}
Consider an FDR problem $\Psi\equiv\langle V,\mathcal{R},I,O,G\rangle$.
Our FDR problem compiled to STRIPS is
{\renewcommand{\theequation}{\roman{equation}}
\setcounter{equation}{0}
 \[
\fdrtostrips (\Psi) \equiv
\left\langle
\begin{gathered}
\fdrtostrips_V \equiv \{\langle v,a\rangle\mid a\in \mathcal{R}(v) \wedge v\in V\},\\
\fdrtostrips_{\mathcal R} \equiv \{\langle v, a\rangle\mapsto \{\top,\bot\} \mid \langle v, a\rangle\in \fdrtostrips_V\},\\
\fdrtostrips_S(I), \fdrtostrips_S(G),\\
\fdrtostrips_O \equiv \{\langle \fdrtostrips_S(p), \fdrtostrips_S(e) \rangle \mid \langle p, e\rangle \in O\}.
\end{gathered}
\right\rangle,\]
}
where, for a (partial) state $s$, $\fdrtostrips_S (s)$ is defined as
{\renewcommand{\theequation}{\roman{equation}}
\setcounter{equation}{0}
 \[
  \begin{multlined}
     \{\langle v,a\rangle \mid s(v)= a \wedge v \in \dom(s)\} \cup \\
     \{\negate{\langle v,a\rangle} \mid s(v)\neq a \wedge v \in \dom(s) \cap V \wedge a \in \mathcal{R}(v)\} 
  \end{multlined}
\]}
\end{myenc}

\begin{myeg}\label{eg:STRIPS}
Consider the FDR problem $\asttofdr(P)$ in example \ref{eg:FDR}.
For that FDR problem, we have
{\renewcommand{\theequation}{\roman{equation}}
\setcounter{equation}{0}
 \[
\left\langle
\begin{gathered}
\fdrtostrips_V\equiv\{\langle v_0, 0\rangle, \langle v_0, 1\rangle\},\\
\fdrtostrips_\mathcal{R}\equiv\{\langle v_0, 0\rangle\mapsto \{\bot, \top\}, \langle v_0, 1\rangle \mapsto \{\bot, \top\}\},\\
\fdrtostrips_S(I)\equiv\{\langle v_0, 0\rangle, \negate{\langle v_0, 1\rangle}\},\fdrtostrips_S(G)\equiv\{\negate{\langle v_0, 0\rangle}, \langle v_0, 1\rangle\},\\
\fdrtostrips_O\equiv\{\langle\{\langle v_0, 0\rangle, \negate{\langle v_0, 1}\rangle\}, \{\negate{\langle v_0, 0\rangle}, \langle v_0, 1\rangle\}\rangle\}
\end{gathered}
\right\rangle
\]
}
\end{myeg}

\begin{mydec}
For an FDR problem $\Psi\equiv\langle V,\mathcal{R},I,O,G\rangle$ and an operator $\langle p,e\rangle\in\fdrtostrips_O$, let 
{\renewcommand{\theequation}{\roman{equation}}
\setcounter{equation}{0}
 \[\stripstofdr_O\langle p,e\rangle\equiv\langle \{v \mapsto a\mid \langle v, a\rangle \in p\}, \{v \mapsto a\mid \langle v, a\rangle \in e\}\rangle\]
}
\end{mydec}
\noindent Note: we ignore negative effects when decoding STRIPS operators since they only ensure operator effect consistency.

The soundness and completeness theorems of this encoding of FDR problems follow.
\begin{mythm}
\label{thm:soundness-strips-fdr-translation}
  For a valid FDR problem $\Psi$ \begin{enumerate*} \item if $[\textit{op}_0, \ldots, \textit{op}_k]$ solves the STRIPS problem $\Pi \equiv \varphi(\Psi)$, then \(\psi \equiv [\varphi_\mathrm O^{-1}(\textit{op}_0), \ldots, \varphi_\mathrm O^{-1}(\textit{op}_k)]\) is a solution for $\Psi$.
\item if $[\textit{op}_0, \ldots, \textit{op}_k]$ solves $\Psi$, then \(\pi \equiv [\varphi_\mathrm O(\textit{op}_0), \ldots, \varphi_\mathrm O(\textit{op}_k)]\)
  solves the STRIPS problem $\varphi(\Pi)$.
\end{enumerate*}
\end{mythm}
\begin{proof}[Proof sketch]
(i) We show that $G\subseteq\exec{I}{\psi}$ and moreover $\varphi_O^{-1}(\textit{op}_0), \ldots, \varphi_O^{-1}(\textit{op}_k)\in\mathcal O$ where $G$ and $\mathcal O$ are the goal state respectively operator set of $\Psi$.
The first part of the proof is by induction over $\pi$ with arbitrary initial state.

\noindent(ii)We show that $G\subseteq\exec{I}{\pi}$ and moreover $\varphi_O(\textit{op}_0), \ldots, \varphi_O(\textit{op}_k)\in\mathcal O$ where $G$ and $\mathcal O$ are the goal state respectively operator set of $\Pi$.
The first part of the proof is by induction over $\psi$ with arbitrary initial state.
\end{proof}

\spacex{
In the proof we first show that given a valid FDR problem $\Psi$, our problem translation $\varphi$ ensures that $\Pi \equiv \varphi(\Psi)$ is also valid STRIPS problem.
\fred{note on why parallel execution semantics are needed}
We then lift the STRIPS solution $\pi$ without conflicting to the trivial corresponding \emph{parallel plan} (i.e. a sequence of operator sets)
  \[\bm\pi \equiv [\{\textit{op}_0\}, \ldots, \{\textit{op}_k\}]\]
For parallel STRIPS solutions with operator sets that are serializable in any order (which in this case is trivially true due to the singleton sets), we prove a lemma which states that
  \[\bm\psi \equiv [\{\varphi_\mathrm O^{-1}(\textit{op}_0)\}, \ldots, \{\varphi_\mathrm O^{-1}(\{\textit{op}_k)\}]\]
is a parallel FDR solution to $\Psi$. 
Lastly, we show that flattening $\bm\psi$ to a serial plan yields a serial solution to $\Psi$.}{} 

\section{Encoding STRIPS Problems as SAT}
\label{sec:stripstosat}
In this step we encode the question of whether a STRIPS problem has a plan of length at most $h$ into a propositional satisfiability formula.
In our formalisation, we use \citeauthor{MichaelisN-TYPES17}'s formalisation of propositional logic.
\fullversion{Due to lack of space and since they are standard, we only describe its syntax and semantics in the appendix.}
The specific encoding we use is similar to the parallel $\forall$-step encoding used by \citeauthor{rintanen:06}.
We limit operators to ones without conditional effects, require a total initial state, and constrain preconditions and the goal to conjunctions of literals.
These restrictions are always satisfied by problems produced by Encoding~\ref{enc:fdrtostrips}.
Informally, such a parallel encoding constitutes an unrolling of the transition relation underlying the STRIPS problem, which allows more than one operator to execute in one time step, as long as those operators are \emph{non-interfering}.
This allows for the encoding to be significantly more compact in practice, compared to only allowing one operator per step.

\newcommand{\intrfr}{\textit{intrfr}}
\newcommand{\add}{\textit{add}}
\newcommand{\del}{\textit{del}}

\begin{mydef}[Interference]
\label{def:interfere}
Two STRIPS operators $o_1\equiv\langle p_1, e_1\rangle$ and $o_2\equiv\langle p_2, e_2\rangle$ are interfering iff $\{v \mid v \mapsto \top\in p_i\} \cap \{v \mid v \mapsto \bot\in e_j\} \neq \emptyset$, for all $i\neq j$, and $i,j \in \{1,2\}$.
For a set of STRIPS operators $O$, we denote the set of pairs of interfering operators in $O$ by $\intrfr(O)$.
\end{mydef}

\newcommand{\stripstosat}{\ensuremath{\Phi}}
\renewcommand{\dots}{..}
\begin{myenc}
\label{enc:forallstep}
Consider a given natural number $h$ and a STRIPS problem $\Pi\equiv(V, R, I, O, G)$.
For a STRIPS state $s$, let $s^t$ denote the propositional formula
{\renewcommand{\theequation}{\roman{equation}}
\setcounter{equation}{0}
 \[(\bigwedge v \in \{v \mid v \mapsto \top \in s\}.\; v^t)\wedge(\bigwedge v \in \{v \mid \negate{v} \in s\}.\; \neg v^t)\]
}
Also, for a variable $v\in V$, let $\add(v)\equiv\{\langle p, e\rangle \mid v \in e\} \cap O$ and $\del(v)\equiv\{\langle p, e\rangle \mid \negate{v} \in e\} \cap O$.
The encoding $\stripstosat(\Pi, h)$ is the conjunction of the following propositional formulae:
{\renewcommand{\theequation}{\roman{equation}}
\setcounter{equation}{0}
 \begin{gather}
 I^0\\
 G^{h}\\
 \begin{multlined} 
 \bigwedge t\in\{0\dots h\}.\; \bigwedge \textit{op} \in O.\; \textit{op}^t\longrightarrow \pre(\textit{op})^t \wedge\\
      \textit{op}^t\longrightarrow \eff(\textit{op})^{t+1}
 \end{multlined}\\
 \begin{multlined} 
 \bigwedge t \in \{1\dots h\}.\; \bigwedge v \in V.\; \neg v^{t - 1} \wedge v^t \longrightarrow\\
                                  \bigvee \textit{op} \in \add(v).\; \textit{op}^t
 \end{multlined}\\
 \begin{multlined} 
 \bigwedge t \in \{1\dots h\}.\; \bigwedge v \in V.\; v^{t - 1} \wedge \neg v^t \longrightarrow\\
  \bigvee \textit{op} \in \del(v).\; \textit{op}^t
 \end{multlined}\\
 \bigwedge t \in \{1\dots h\}.\; \bigwedge \langle \textit{op}, \textit{op}'\rangle\in \intrfr(O).\; \textit{op}^t \vee \neg \textit{op}'^t
\end{gather}}
This encoding is defined over the atoms 
{\renewcommand{\theequation}{\roman{equation}}
\setcounter{equation}{0}
 \[\{v^t \mid v \in V \wedge 0 \leq t \leq h\} \cup \{\textit{op}^t \mid \textit{op} \in O \wedge 0 \leq t < h\}\]
}
\end{myenc}

In the encoding above the first conjunct stands for the initial state, the second for the goal, the third for the transition relation, the fourth and fifth are the frame axioms, and the last is a constraint ensuring that if more than one operator execute in the same step, they are not interfering operators.
Also note that the actual encoding we verified only computes the formula in CNF form, but we use syntactic sugar in our definition and examples to improve readability, e.g.\ $x_1 \wedge x_2 \longrightarrow \bigvee y \in \{y_1, y_2, \dots\}. y$ is syntactic sugar for $\neg x_1 \vee \neg x_2 \vee y_1 \vee y_2 \dots$, and $x \longrightarrow \bigwedge y \in \{y_1, y_2, \dots\}. y$ is syntactic sugar for $(\neg x \vee y_1) \wedge (\neg x \vee y_2) \wedge \dots$.

\begin{myeg}
\label{eg:SAT}
Consider the STRIPS problem $\fdrtostrips(\asttofdr(P))$ from Example~\ref{eg:STRIPS}.
Let $h \equiv 1$ be the horizon.
The encoding is the conjunction of 
{\renewcommand{\theequation}{\roman{equation}}
\setcounter{equation}{0}
 \begin{gather}
\langle v_0, 0\rangle^0 \wedge \neg\langle v_0, 1\rangle^0 \\
\neg\langle v_0, 0\rangle^1 \wedge \langle v_0, 1\rangle^1 \\
\begin{multlined} 
  (\textit{op}_0^0\longrightarrow \langle v_0, 0\rangle^0 \wedge \neg\langle v_0, 1\rangle^0) \wedge\\
  (\textit{op}_0^0\longrightarrow \neg\langle v_0, 0\rangle^1 \wedge \langle v_0, 1\rangle^1)
\end{multlined}\\
\begin{multlined} 
(\neg \langle v_0, 0\rangle^0 \wedge \langle v_0, 0\rangle^1 \longrightarrow \bot)\wedge\\
(\neg \langle v_0, 1\rangle^0 \wedge \langle v_0, 1\rangle^1 \longrightarrow \textit{op}_0^0)
\end{multlined}\\
\begin{multlined} 
(\langle v_0, 0\rangle^0 \wedge \neg\langle v_0, 0\rangle^1 \longrightarrow \textit{op}_0^0) \wedge\\
(\langle v_0, 1\rangle^0 \wedge \neg\langle v_0, 1\rangle^1 \longrightarrow \bot)
\end{multlined}\\
 \top
\end{gather}}
\end{myeg}

\newcommand{\listing}{\textit{list}}
\newcommand{\flatten}{\textit{flat}}
\newcommand{\model}{\textit{flat}}
\newcommand{\modeltostrips}{\ensuremath{\Phi^{-1}}}

\begin{mydec}
\label{dec:forallstep}
Consider a horizon $h$, a STRIPS problem $\Pi\equiv(V, R, I, O, G)$ and a model $\mathcal{M} \vDash \stripstosat(\Pi, h)$.
Let, for a set $s$, $\listing(s)$ denote an arbitrary list which contains all the elements of $s$, s.t.\ $\cardinality{s}=\cardinality{\listing(s)}$.
Let for a list of lists $\textit{ls}\equiv[l_0,l_1,\dots]$, $\flatten(\textit{ls})$ denote the list $l_0\cat l_1, \dots$, where $\cat$ is the list append function.
The decoding function is defined $\modeltostrips(\Pi, h, \mathcal{M})\equiv\flatten([\listing(\{\textit{op} \mid \textit{op}^0 \in \mathcal{M}\}), \listing(\{\textit{op} \mid \textit{op}^1 \in \mathcal{M}\}),\dots, \listing(\{\textit{op} \mid \textit{op}^{h-1} \in \mathcal{M}\})])$.
\end{mydec}

\begin{myeg}\label{eg:decoding}
  For the propositional formula in Example~\ref{eg:SAT}, a model is $\{\langle v_0, 0\rangle^0, \negate{\langle v_0, 1\rangle^0}, \negate{\langle v_0, 0\rangle^1}, \langle v_0, 1\rangle^1, \textit{op}_0^0\}$. 
  The decoded plan is $[\textit{op}_0]$.
\end{myeg}

This translation to SAT is sound and complete.
\begin{mythm}
\label{thm:toSAT}
For a valid STRIPS problem $\Pi$ and a horizon $h$:\begin{enumerate*}\item if $\mathcal M$ is a model for $\Phi(\Pi, h)$, then $\modeltostrips(\Pi, h,\mathcal M)$ is a solution for $\Pi$ 
  and \item if $\fdrops$ is a solution for $\Pi$ and $\cardinality{\fdrops}\leq h$, then there is a model for $\Phi(\Pi, h)$.\end{enumerate*}
\end{mythm}
\begin{proof}[Proof sketch]
(i) This proof is by induction on the horizon, with generalising the initial state.
It depends, crucially, on the fact that non-interference implies that any order of operators coming from a single step is executable.

\noindent(ii) Let the operators in $\textit{ops}$ be $[\textit{op}_0,\textit{op}_1,\dots,\textit{op}_{\cardinality{\as}-1}]$.
We construct a model for $\Phi(\Pi, h)$ by considering the sequence of states traversed by executing the plan $\pi$ at $I$, which can be recursively specified as $s_0 \equiv I$, and $s_{t+1}\equiv\exec{s_t}{\textit{op}_t}$ for $0 < t \leq h$.
The model we construct is 
{\renewcommand{\theequation}{\roman{equation}}
\setcounter{equation}{0}
 \begin{gather*}
\begin{multlined}
\{\textit{op}_t^t \mid 0\leq t < \cardinality{\as}\} \cup\\
\{\negate{\textit{op}_t^{t'}} \mid 0\leq t\neq t' < \cardinality{\as}\}\cup \{s_t^t \mid 0\leq t \leq \cardinality{\as}\}
\end{multlined}
\end{gather*}}
\end{proof}

Before we conclude, we note that other encoding methods from FDR to SAT have also been proposed, e.g.\ \citeauthor{balyo2013relaxing}~\citeyear{balyo2013relaxing}.

\section{Abstract SAT Formulae to DIMACS}
\label{sec:sattodimacs}

The SAT formulae produced by Encoding~\ref{enc:forallstep} are structured in the following way: \begin{enumerate*}\item the formulae use the connectives $\wedge, \vee, \rightarrow$ and $\neg$ and \item atoms representing state variables and operators are indexed by the time step.\end{enumerate*}
In a pen-and-paper exposition this would be enough.
However, this is not enough in our case because there is still an encoding step to simplify these structured formulae to DIMACS-CNF, which is the representation of SAT formulae used in practice, and we do not want to trust that step.
Thus, as a last step, we present the following encoding of structured formulae to DIMACS-CNF ASTs.
That encoding has to simplify the connectives as well as replace the structured variables with integers.
\begin{mydef}[DIMACS-CNF]
\label{def:dimacs}
A DIMACS-CNF AST is a list of lists of non-zero integers.
A list of non-zero integers $[l_1,l_2,\dots,l_m]$ is a model for a DIMACS-CNF AST $[c_1\equiv[l_{11},l_{12}\dots],c_2\equiv[l_{21},l_{22}\dots],\dots,c_n\equiv[l_{n1},l_{n2}\dots]]$ iff for each $c_i$, for $1 \leq i \leq n$, there is $l_j\in c_i$, where $1 \leq j \leq m$, and, for each $1\leq j,k\leq m$, $l_j \neq - l_k$.
\end{mydef}
\newcommand{\simp}{\ensuremath{\textit{simp}}}
\newcommand{\simpdisj}{\ensuremath{\textit{simpORs}}}
\newcommand{\intofvar}{\ensuremath{\textit{int}}}
\newcommand{\stripsop}{\ensuremath{\textit{op}}}
\newcommand{\varofint}{\ensuremath{\textit{var}}}
\newcommand{\litofint}{\ensuremath{\textit{lit}}}
\newcommand{\sign}{\ensuremath{\textit{sign}}}

\begin{myenc}
\label{enc:toDIMACS}
For a STRIPS problems $(V,R,I,O,G)$, fix an arbitrary ordering $\bf V$ ($\bf O$) of $V$ ($O$) s.t.\ for a state variable (operator) $v$ ($\stripsop$), $v_i$ ($\stripsop_i$) is the index of $v$ ($\stripsop$) in ${\bf V}$ (${\bf O}$).
For an atom $a$ and a horizon $h$, let $\intofvar(a)$ be defined as: \begin{enumerate*}\item $1+t+\stripsop_i(h+1)$, if $\exists\stripsop^t.\; a = \stripsop^t$ and \item $1 + \cardinality{O}(h+1) + t + v_i(h+1)$ if $\exists v^t.\; a = v^t$\end{enumerate*}.
For a propositional formula $\phi$ that has no conjunctions, let $\simpdisj(\phi)$ be defined recursively as follows:\begin{enumerate*}\item $\simpdisj(\phi_1)\cat\simpdisj(\phi_2)$, if $\exists\phi_1,\phi_2.\; \phi= \phi_1\vee\phi_2$, \item $[]$, if $\phi=\bot$, \item $[-1,1]$, if $\phi=\top$, \item $[-\simp(\phi')]$, if $\exists\phi'.\; \phi= \neg\phi'$, and \item $[\intofvar(\phi)]$, if $\phi$ is an atom.\end{enumerate*}
Let $\simp(\phi)$ be defined recursively as follows:\begin{enumerate*}\item $\simp(\phi_1) \cat \simp(\phi_2)$, if $\exists\phi_1,\phi_2.\; \phi= \phi_1\wedge\phi_2$ and \item $[\simpdisj(\phi)]$ otherwise.\end{enumerate*}
\end{myenc}
\noindent Note: $\intofvar$ and $\varofint$ are adapted from \citeauthor{DBLP:books/lib/Knuth98}~\citeyear[Section 4.4]{{DBLP:books/lib/Knuth98}} on encoding numbers with arbitrary radixes.
Also, note that $\simp$ is only well-defined if the given formula is a CNF, which is not problematic since the formulae produced by $\Phi$ are CNF for a valid STRIPS problem.
\begin{myeg}
\label{eg:DIMACS}
\makeatletter
\def\old@comma{,}
\catcode`\,=13
\def,{\ifmmode\old@comma\discretionary{}{}{}\else\old@comma\fi}
\makeatother
\makeatletter
\def\old@dot{.}
\catcode`\.=13
\def.{\ifmmode\old@dot\discretionary{}{}{}\else\old@dot\fi}
\makeatother
 Consider the CNF formula from Example~\ref{eg:SAT}.
First consider the following two orderings for the variables encoding operators and state variables: $[\textit{op}_0]$ and $[\langle v_0, 0\rangle,\langle v_0, 1\rangle]$.
The mapping of the time indexed variables to natural numbers is: $\intofvar(\textit{op}_0^0)=1$, $\intofvar(\textit{op}_0^1)=2$, $\intofvar(\langle v_0, 0\rangle^0)= 3$, $\intofvar(\langle v_0, 0\rangle^1)=4$, $\intofvar(\langle v_0, 1\rangle^0)=5$, and $\intofvar(\langle v_0, 1\rangle^1)=6$.
Applying $\simp$ to that formula results in $[[3],[-5],[-4],[6],[-1,3],[-1,-5],[-1,-4],[-1,6],[3,-4],[5,-6,1],[-3,4,1],[-5,6],[-1,1]]$.
This AST is then pretty printed as DIMACS-CNF concrete syntax, like the one in Fig.~\ref{fig:concretesyntax}, which only shows 7 clauses.
\end{myeg}
\begin{mydec}
\label{dec:toDIMACS}
Consider a DIMACS-CNF AST encoding a STRIPS problem $(V,R,I,O,G)$ and a horizon $h$.
Let, for an integer $n$, the function $\varofint(n)$ be defined as: \begin{enumerate*}\item $({\bf O}(\cardinality{n} - 1 \mod h))^{(\cardinality{n} - 1)\div h}$, if $\cardinality{n} < 1 + h\cardinality{O}$ \item $({\bf V}(k \mod h))^{k \div h}$, where $k = \cardinality{n} - h\cardinality{O} - 1$, otherwise.
Now, let $\litofint(n)$ be the literal $\varofint(n)$, if $0 < n$, and $\neg\varofint(n)$ otherwise.
\end{enumerate*}
\end{mydec}
\noindent The correctness theorem for this step is as follows.
\begin{mythm}
\label{thm:toDIMACS}
For a valid STRIPS problems $\Pi$ and a horizon $h$: \begin{enumerate*}\item if $\{l_1, l_2, \dots\}$ is a model for $\Phi(\Pi, h)$, then $\{\simp(l_1), \simp(l_2), \dots\}$ is a model for $\simp(\Phi(\Pi, h))$, and \item if $[n_1, n_2, \dots]$ is a model for $\simp(\Phi(\Pi, h))$, then $\{\litofint(n_1), \litofint(n_2), \dots\}$ is a model for $\Phi(\Pi, h)$.\end{enumerate*}
\end{mythm}
\begin{proof}[Proof sketch]
The proof of both statements is by structural induction on the formula $\Phi(\Pi, h)$.
\end{proof}

\fullversion{The overall encoding correctness follows (Listing~\ref{isa:finamThm})}.
\begin{mythm}
\label{thm:soundAndComplete}
For a valid FD-AST $P$ and a horizon $h$: \begin{enumerate*}\item if $[n_1, n_2, \dots]$ is a model for $\simp(\Phi(\fdrtostrips(\asttofdr(P)), h))$ then a plan for $P$ is $[\fdrtoast(\stripstofdr_O(\fdrop_1)),\fdrtoast(\stripstofdr_O(\fdrop_2)),\dots]$, where $[\fdrop_1,\fdrop_2,\dots] = \Phi^{-1}(\Pi, h, \{\litofint(n_1), \litofint(n_2), \dots\})$, and \item if $[\opname_1, \opname_2, \dots, \opname_h]$ is a plan for $P$, then there is a model for  $\simp(\Phi(\varphi(\asttofdr(P)), h))$.\end{enumerate*}
\end{mythm}
\begin{proof}[Proof sketch]
From Theorems~\ref{thm:asttofdr}, \ref{thm:soundness-strips-fdr-translation}, \ref{thm:toSAT}~and \ref{thm:toDIMACS}.
\end{proof}

 \begin{table}[t]
\centering
    \begin{tabularx}{0.45\textwidth}{ c  c  c  c  c }
          \phantom{abc}       &\multicolumn{2}{c}{\tiny SAT}              &\multicolumn{2}{c}{\tiny UNSAT}       \\ 
    \hline
          {\tiny }       &{\tiny Madagascar}       &{\tiny Verified Encoding}      &{\tiny Madagascar}       &{\tiny Verified Encoding}\vspace{-0.5ex}\\ 
{\scriptsize newopen}       &{\scriptsize 3128}       &{\scriptsize 2897}       &{\scriptsize 2214}       &{\scriptsize 2205}\\  
{\scriptsize logistics}       &{\scriptsize 986}       &{\scriptsize 358}       &{\scriptsize 1031}       &{\scriptsize 452}\\  
{\scriptsize elevators}       &{\scriptsize 75}       &{\scriptsize 14}       &{\scriptsize 61}       &{\scriptsize 44}\\  
{\scriptsize rover}       &{\scriptsize 270}       &{\scriptsize 233}       &{\scriptsize 172}       &{\scriptsize 121}\\  
{\scriptsize storage}       &{\scriptsize 66}       &{\scriptsize 35}       &{\scriptsize 41}       &{\scriptsize 33}\\  
{\scriptsize pipesworld}       &{\scriptsize 46}       &{\scriptsize 9}       &{\scriptsize 74}       &{\scriptsize 7}\\
{\scriptsize nomystery}       &{\scriptsize 83}       &{\scriptsize 13}       &{\scriptsize 197}       &{\scriptsize 15}\\  
{\scriptsize zeno}       &{\scriptsize 180}       &{\scriptsize 54}       &{\scriptsize 70}       &{\scriptsize 28}\\  
{\scriptsize hiking}       &{\scriptsize 45}       &{\scriptsize 6}       &{\scriptsize 100}       &{\scriptsize 11}\\  
{\scriptsize TPP}       &{\scriptsize 106}       &{\scriptsize 46}       &{\scriptsize 71}       &{\scriptsize 30}\\  
{\scriptsize Transport}       &{\scriptsize 78}       &{\scriptsize 2}       &{\scriptsize 134}       &{\scriptsize 18}\\  
{\scriptsize GED}       &{\scriptsize 68}       &{\scriptsize 19}       &{\scriptsize 105}       &{\scriptsize 19}\\  
{\scriptsize woodworking}       &{\scriptsize 99}       &{\scriptsize 51}       &{\scriptsize 17}       &{\scriptsize 20}\\  
{\scriptsize visitall}       &{\scriptsize 74}       &{\scriptsize 53}       &{\scriptsize 279}       &{\scriptsize 131}\\  
{\scriptsize openstacks}       &{\scriptsize 62}       &{\scriptsize --- }       &{\scriptsize 316}       &{\scriptsize 62}\\  
{\scriptsize satellite}       &{\scriptsize 46}       &{\scriptsize 34}       &{\scriptsize 20}       &{\scriptsize 20}\\  
{\scriptsize scanalyzer}       &{\scriptsize 88}       &{\scriptsize 4}       &{\scriptsize 88}       &{\scriptsize 9}\\  
{\scriptsize tidybot}       &{\scriptsize 24}       &{\scriptsize --- }       &{\scriptsize 12}       &{\scriptsize --- }\\  
{\scriptsize trucks}       &{\scriptsize 45}       &{\scriptsize 8}       &{\scriptsize 117}       &{\scriptsize 29}\\  
{\scriptsize parcprinter}       &{\scriptsize 113}       &{\scriptsize 63}       &{\scriptsize 128}       &{\scriptsize 111}\\  
{\scriptsize maintenance}       &{\scriptsize 35}       &{\scriptsize 34}       &{\scriptsize --- }       &{\scriptsize --- }\\  
{\scriptsize pegsol}       &{\scriptsize 114}       &{\scriptsize 3}       &{\scriptsize 200}       &{\scriptsize 237}\\  
{\scriptsize blocksworld}       &{\scriptsize 31}       &{\scriptsize 25}       &{\scriptsize 30}       &{\scriptsize 30}\\  
{\scriptsize floortile}       &{\scriptsize 221}       &{\scriptsize 54}       &{\scriptsize 361}       &{\scriptsize 281}\\  
{\scriptsize barman}       &{\scriptsize 20}       &{\scriptsize --- }       &{\scriptsize 208}       &{\scriptsize 34}\\  
{\scriptsize Thoughtful}       &{\scriptsize --- }       &{\scriptsize --- }       &{\scriptsize --- }       &{\scriptsize 5}\\  

    \hline
\end{tabularx}
\caption{\label{table:solved} Number of solved satisfiable and unsatisfiable formulae solved by Kissat for our encoding and the encoding generated by Madagascar.}
\end{table}
 \section{Experimental Evaluation}
\label{sec:experiments}

We use Isabelle/HOL's code generator to generate a Standard ML implementation of our correct encoding.
Readers interested in implementation details can inspect the attachement.
We evaluate the performance of our encoding compared to the $\forall$-step encoding of \citeauthor{rintanen:06}~\citeyear{rintanen:06} as computed by Madagascar when invariant generation is disabled.
Although this setup has a weakness, namely, that Madagascar takes the PDDL domain as input while our system takes the grounded output of Fast Downward's translator, it should indicate the scalability of the verified encoding and can be used to test Madagascar's completeness.
We compute the encodings of different planning domains from previous competitions and then feed them to the SAT solver Kissat, opting for a 30 minutes timeout and a 8GB memory limit for encoding and solving.
We generate the encodings for horizons 2, 5, 10, 20, 50, 100, and the bounds generated by the algorithm of \citeauthor{aaai2019}~\citeyear{aaai2019}.
We record a few findings.
Firstly, we found a bug in Madagascar: it produces unsatisfiable formulae for instances and horizons, despite those instances having $\forall$-step plans bounded by the horizon.
This happens in at least 24 instances of different variants of the Rovers and PARCPrinter domains.
This is because Madagascar adds incorrect action mutex constraints which rule out valid $\forall$-step plans, thus causing Madagascar to not be complete.
The fact that such a well-established planning system has such bugs demonstrate that it is imperative we verify planning systems, especially that there no generally succinct unsolvablity certificates do not exist for AI planning algorithms.

Secondly, we compare the performances of our encoding and Madagascar.
\begin{enumerate*}
\item For most instances, our encoding is solved by Kissat in significantly shorter time than Madagascar\fullversion{ (see Fig.~\ref{fig:allKissatTime})}.
Kissat fails to terminate on Madagascar encodings of some Rovers instances, while it succeeds for our encodings.
We note that Madagascar's $\forall$-step encoding is linear in size due to the use of auxiliary variables to represent the operator interference clauses, while ours is quadratic\fullversion{ (see Fig.~\ref{fig:allClauses})}.
A hypothesis is that auxiliary variables interfere with Kissat's deduction mechanisms, as has been reported about compact encodings in other contexts~\cite{knuth2015art}[Section 7.2.2.2].
However, this needs further study.
\item
Since our verified implementation is purely functional in Standard ML, computing our encoding takes longer time than Madagascar's, e.g.\ we use balanced trees instead of arrays, causing every access/update to be worst-case logarithmic instead of constant time.
This leads to our encoding to have a worse total (i.e.\ grounding, encoding and solving) running time, despite the fact that our encoding is usually solvable in shorter time\fullversion{ (see Fig.~\ref{fig:allTotalTime})}.
A bigger problem is that, as Standard ML does not support lazy evaluation and has poor memory management in general, our encoding frequently runs out of memory as it computes the entire encoding in memory before producing any output\fullversion{ (see Fig.~\ref{fig:allTotalMemory})}.
This leads to less of our encodings being solved by Kissat compared to the ones produced by Madagascar (see Table~\ref{table:solved}).

\end{enumerate*}
 \vspace{-0.5ex}
\section{Discussion}
\vspace{-0.5ex}
\label{sec:conclusion}
We presented an executable formally verified SAT encoding of AI planning.
We showed details of the verification process, and experimentally tested our encoding.
Experiments show that, although our verified encoder is primarily hindered by its memory consumption, it can handle planning problems of reasonable sizes, where it can solve, or show bounded length plan non-existence. 
By testing Madagascar's encoding against our verified encoding, we discovered that Madagascar sometimes mistakenly claims that problems have no solutions of a certain length.
Also, compared to Madagascar, our encoding can be more efficiently processed by the SAT solver Kissat.
The size of the verified Standard ML program is around 1.2K lines of code, and the size of the formal proof is around 17.5K lines of proof scripts.

One goal of our work here is to showcase theorem proving and its application to verification as a methodology to increase trustworthiness of planning software and, more generally, AI software.
Although there are other approaches to increase reliability of AI systems, most notably certification for planning or SMT-based methods for verifying properties of neural networks, we believe that correct-by-construction algorithms have their niche.
For instance, this is the case when there are not general certification methods or when desired formal properties are too complex for automated methods to practically handle.

As future work, the most interesting direction is to verify the encoding of costs by~\citeauthor{DBLP:journals/corr/abs-2103-02355}~\citeyear{DBLP:journals/corr/abs-2103-02355}, yielding formally verified certificates of cost optimality.
Another direction is optimising the memory consumption of our implementation, e.g.\ via lazy evaluation~\cite{lochbihler2018lazy}, or by using a low-level target language instead of Standard ML, like LLVM~\cite{LammichLLVM}.

 \bibliography{short_paper}
\clearpage
\appendix
\renewcommand\thefigure{\thesection.\arabic{figure}}
\renewcommand\thetable{\thesection.\arabic{table}}

\begin{figure*}[h]
\centering
\begin{minipage}[b]{0.24\textwidth}
\centering
        \includegraphics[width=1\textwidth,height=0.8\textwidth]{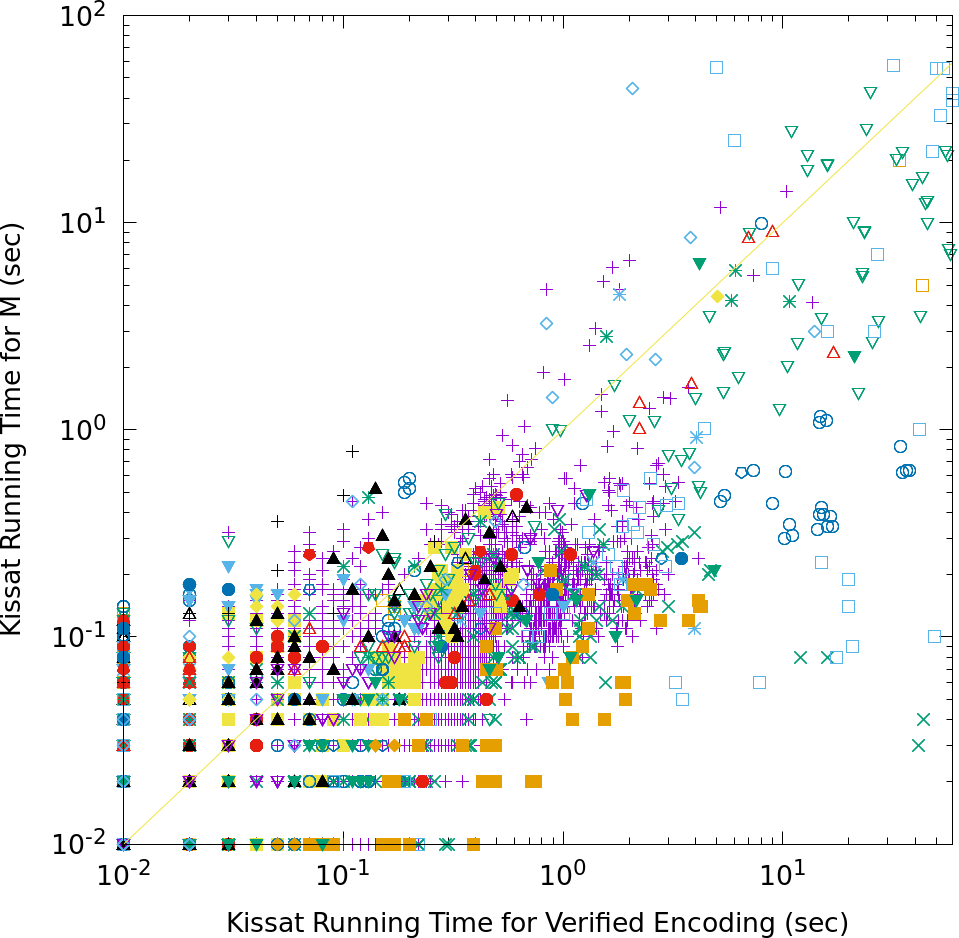}
\caption{\label{fig:allKissatTime}}
\end{minipage}
\begin{minipage}[b]{0.24\textwidth}
\centering
        \includegraphics[width=1\textwidth,height=0.8\textwidth]{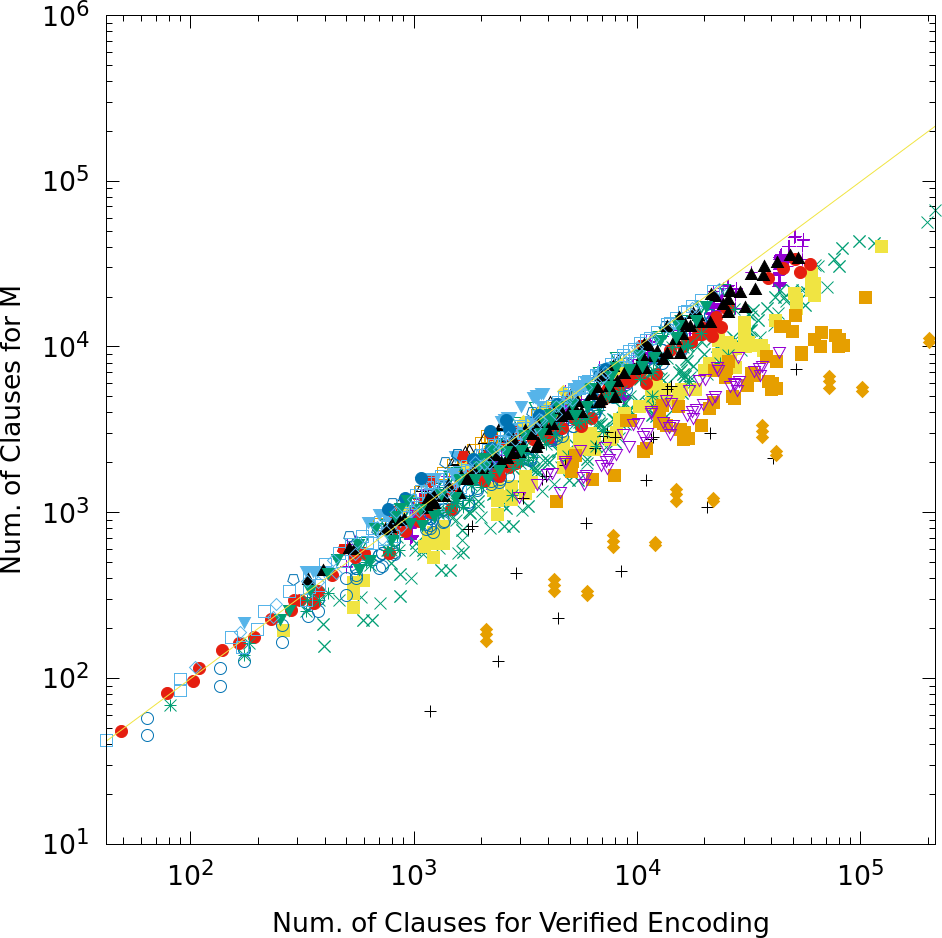}
\caption{\label{fig:allClauses}}
\end{minipage}
\begin{minipage}[b]{0.24\textwidth}
\centering
        \includegraphics[width=1\textwidth,height=0.8\textwidth]{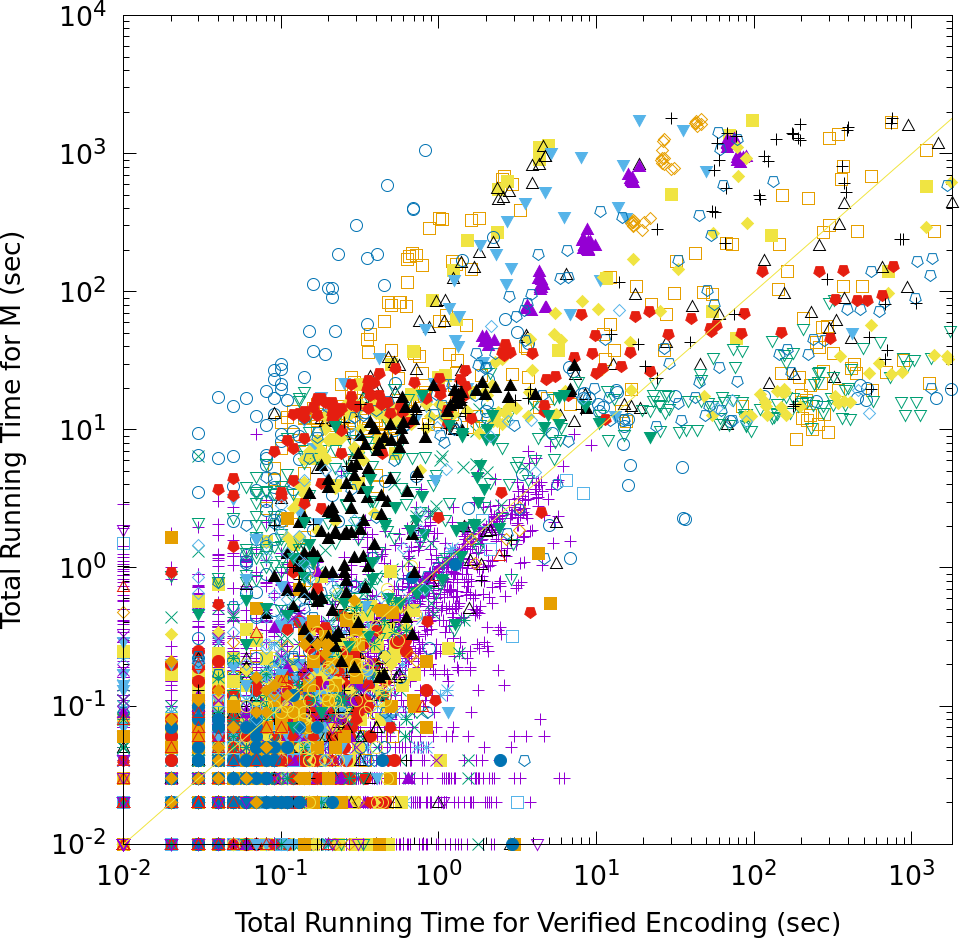}
\caption{\label{fig:allTotalTime}}
\end{minipage}
\begin{minipage}[b]{0.24\textwidth}
\centering
        \includegraphics[width=1\textwidth,height=0.8\textwidth]{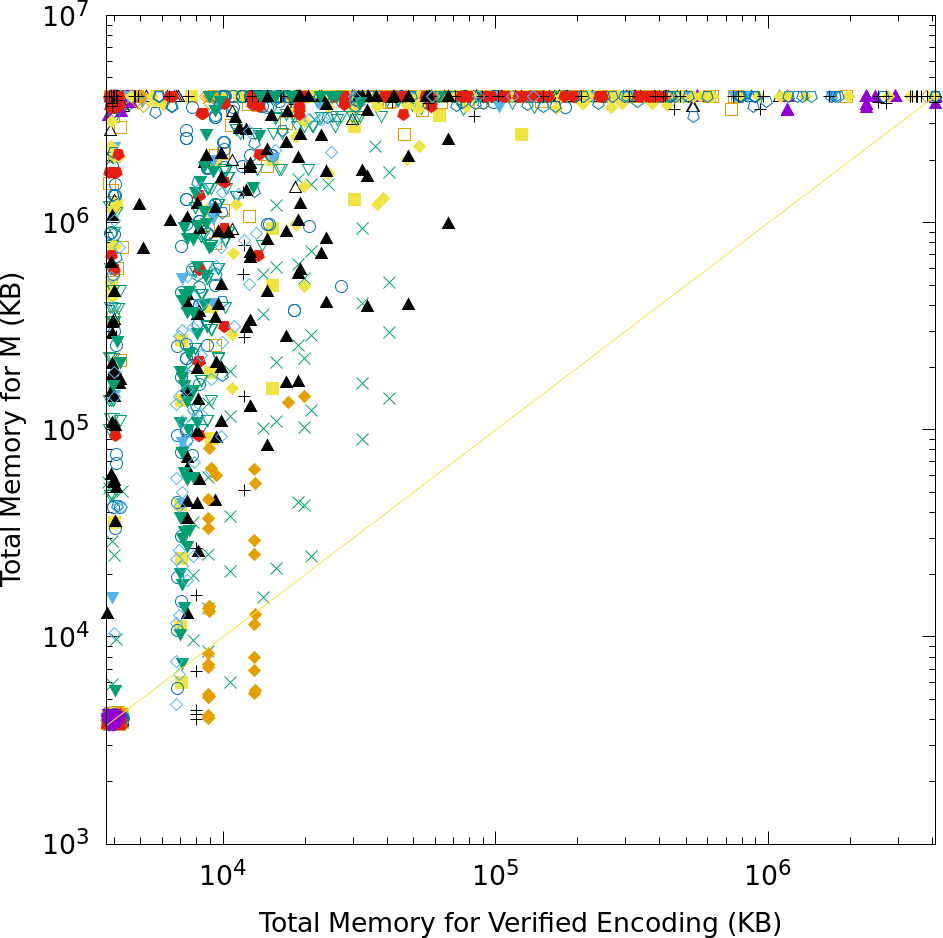}
\caption{\label{fig:allTotalMemory}}
\end{minipage}
\begin{minipage}[b]{0.24\textwidth}
\centering
       \includegraphics[width=1\textwidth,height=0.8\textwidth]{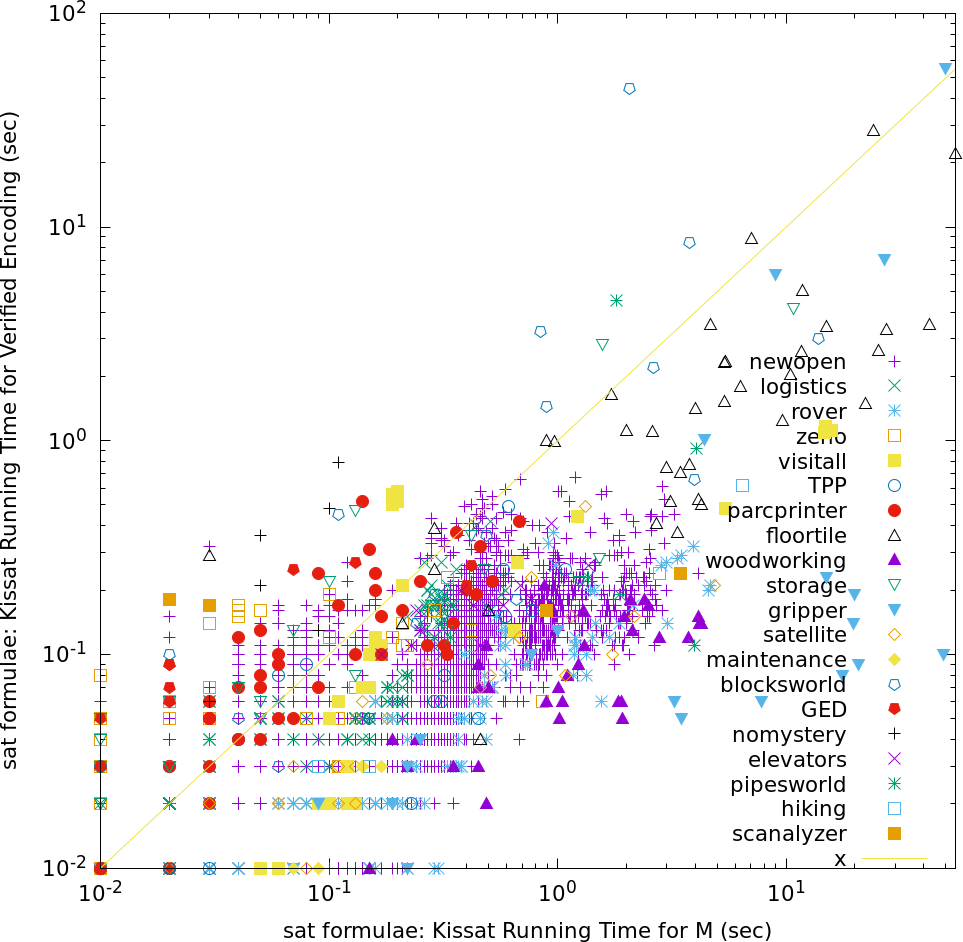}
\end{minipage}
\begin{minipage}[b]{0.24\textwidth}
\centering
        \includegraphics[width=1\textwidth,height=0.8\textwidth]{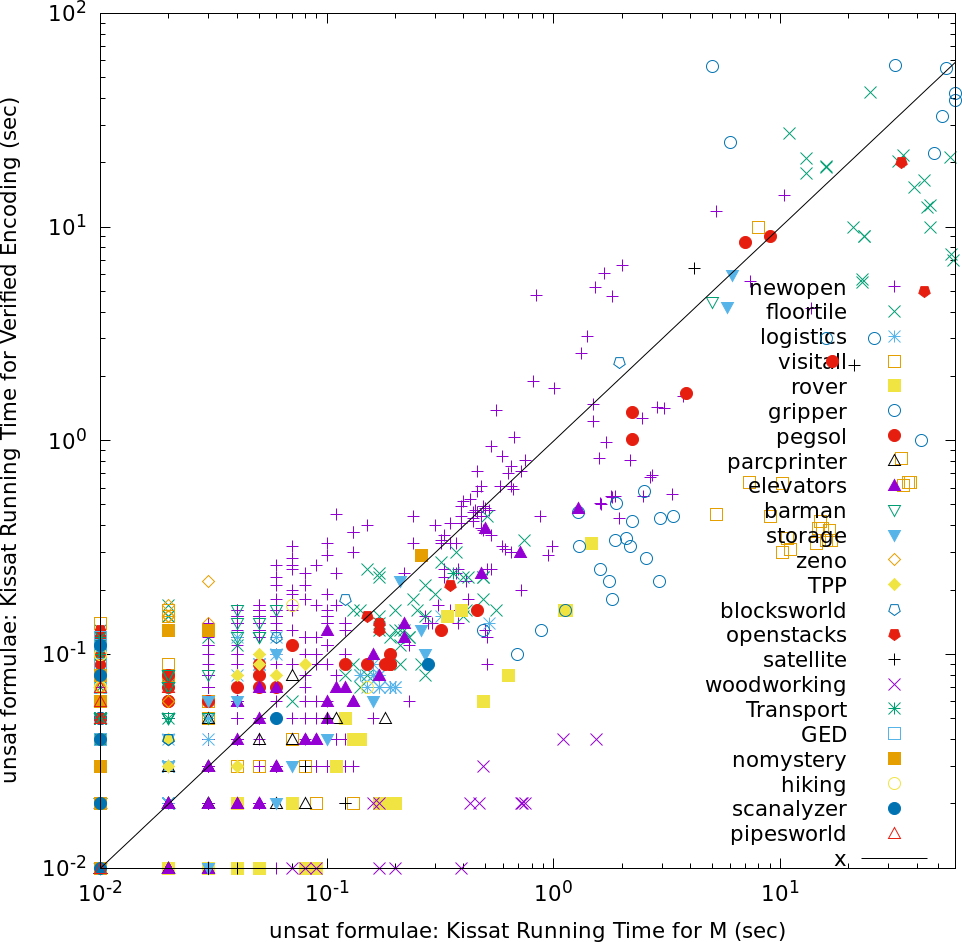}
\end{minipage}
\begin{minipage}[b]{0.24\textwidth}
\centering
        \includegraphics[width=1\textwidth,height=0.8\textwidth]{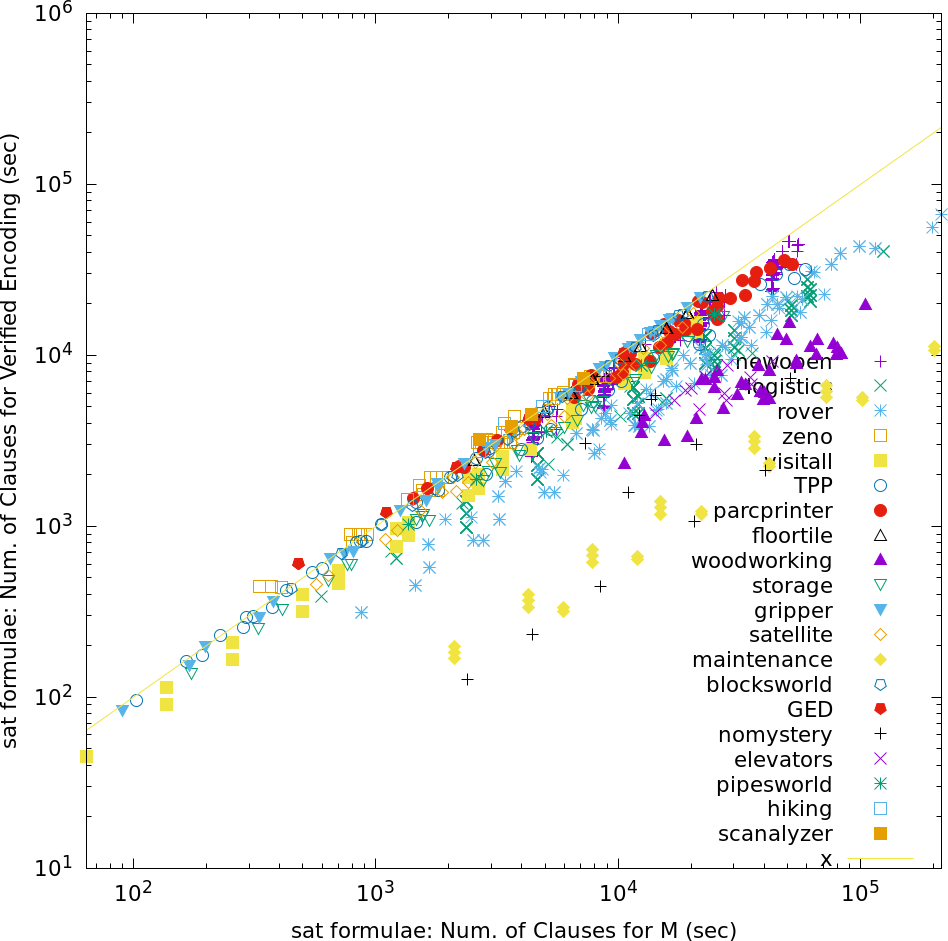}
\end{minipage}
\begin{minipage}[b]{0.24\textwidth}
\centering
        \includegraphics[width=1\textwidth,height=0.8\textwidth]{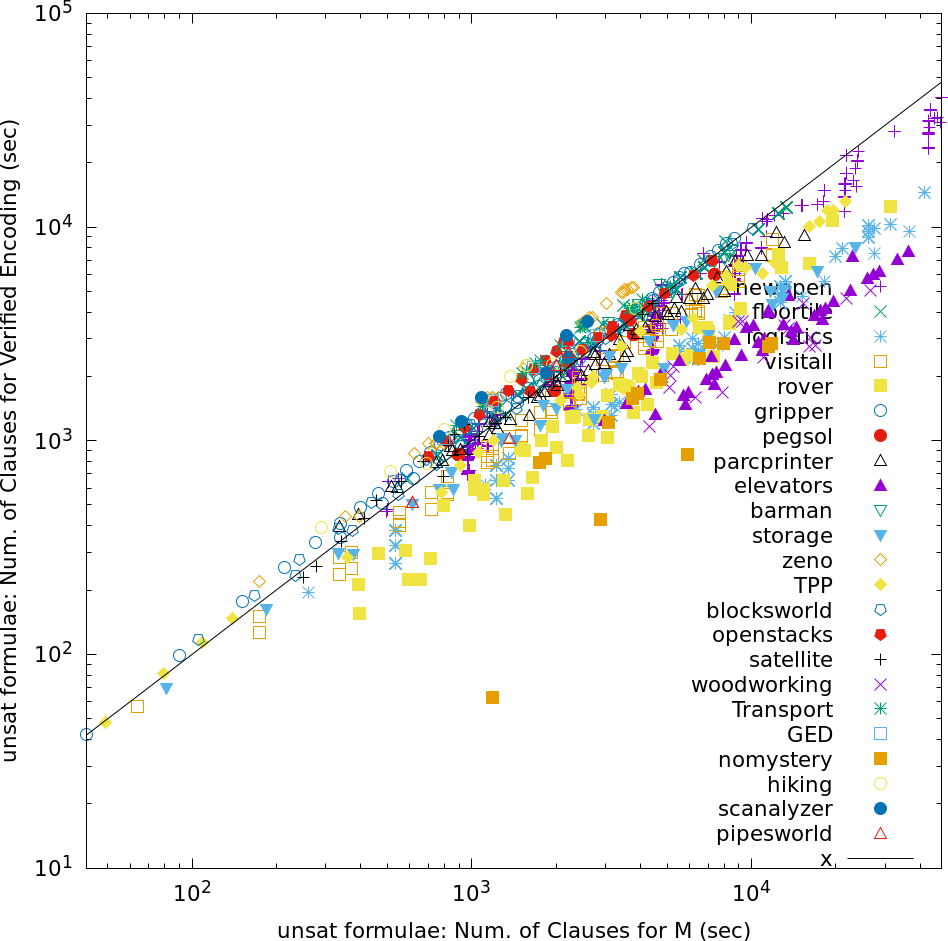}
\end{minipage}
\begin{minipage}[b]{0.24\textwidth}
\centering
        \includegraphics[width=1\textwidth,height=0.8\textwidth]{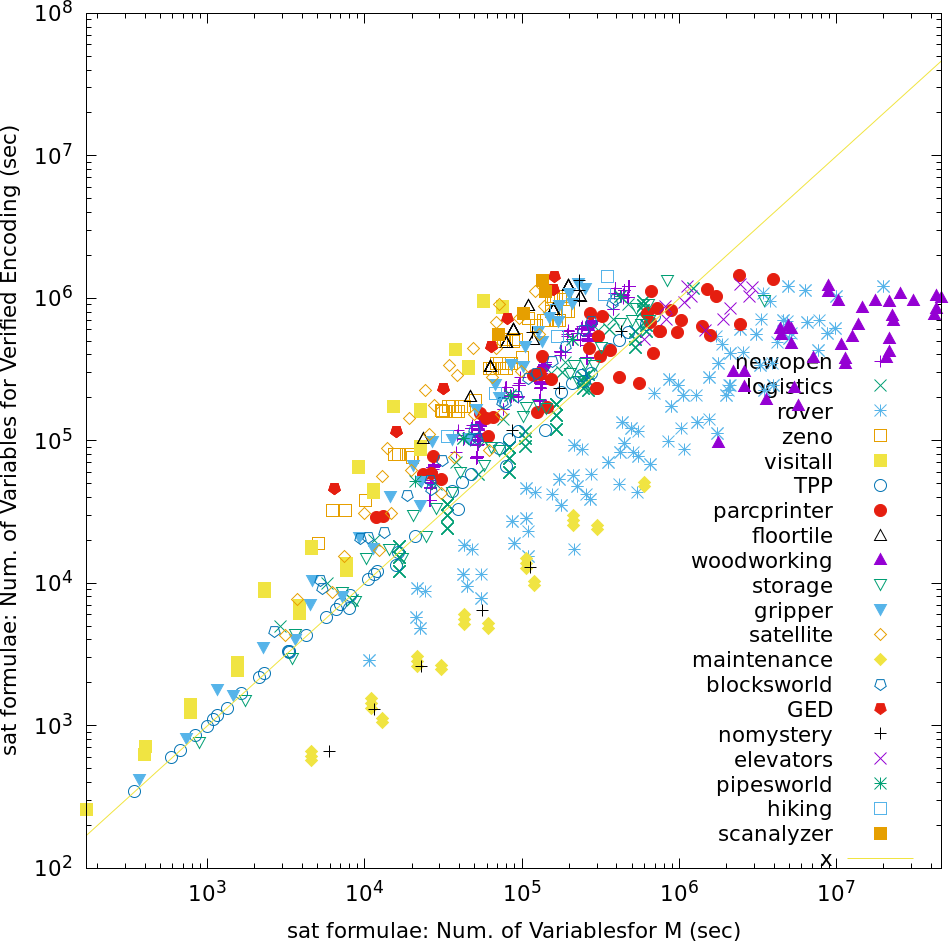}
\end{minipage}
\begin{minipage}[b]{0.24\textwidth}
\centering
        \includegraphics[width=1\textwidth,height=0.8\textwidth]{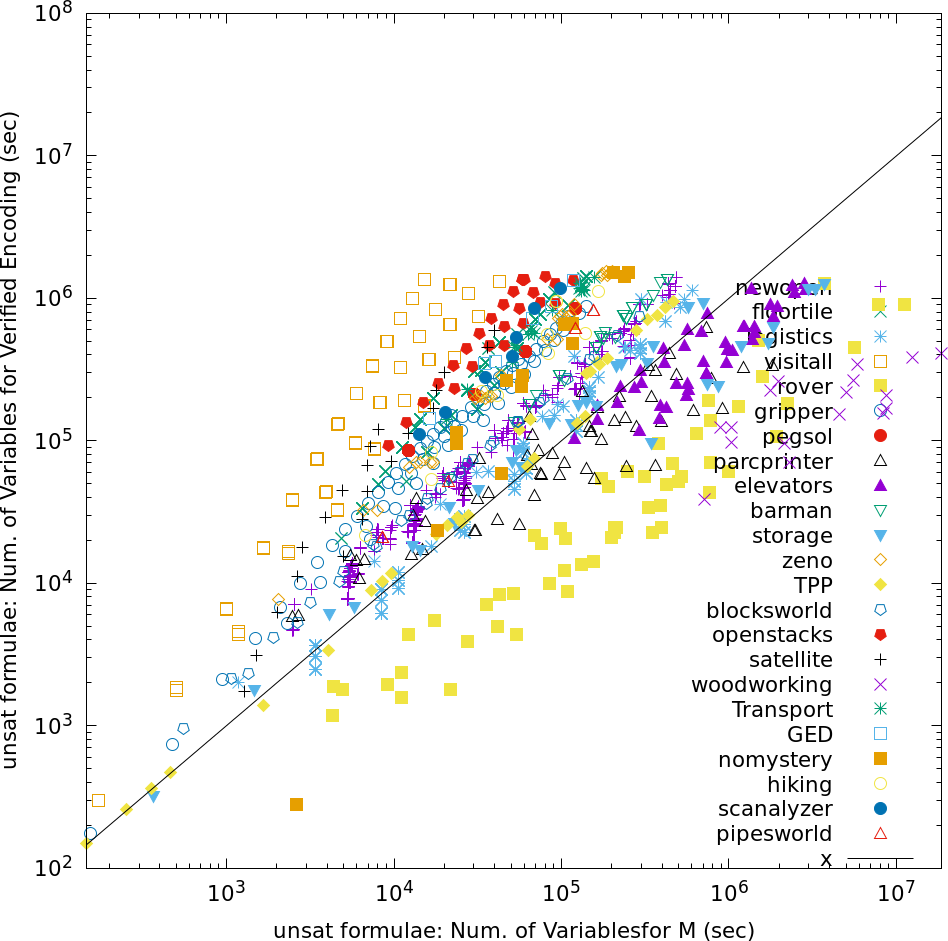}
\end{minipage}
\begin{minipage}[b]{0.24\textwidth}
\centering
        \includegraphics[width=1\textwidth,height=0.8\textwidth]{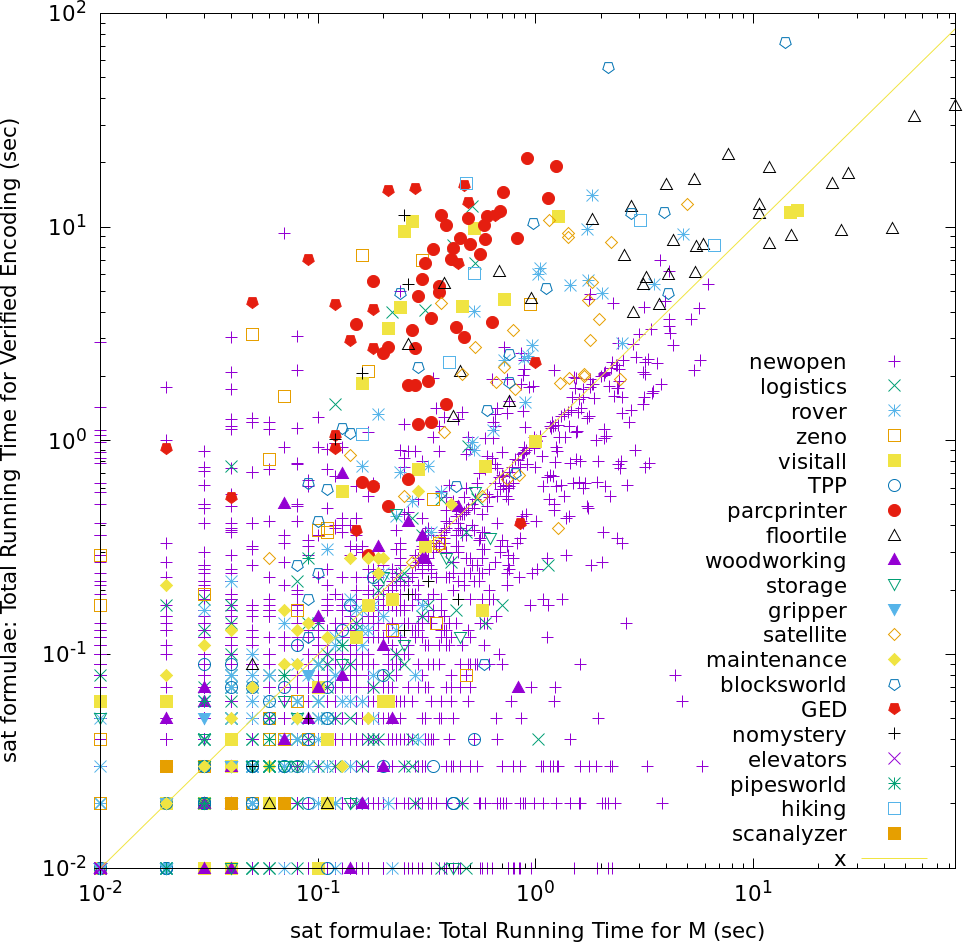}
\end{minipage}
\begin{minipage}[b]{0.24\textwidth}
\centering
        \includegraphics[width=1\textwidth,height=0.8\textwidth]{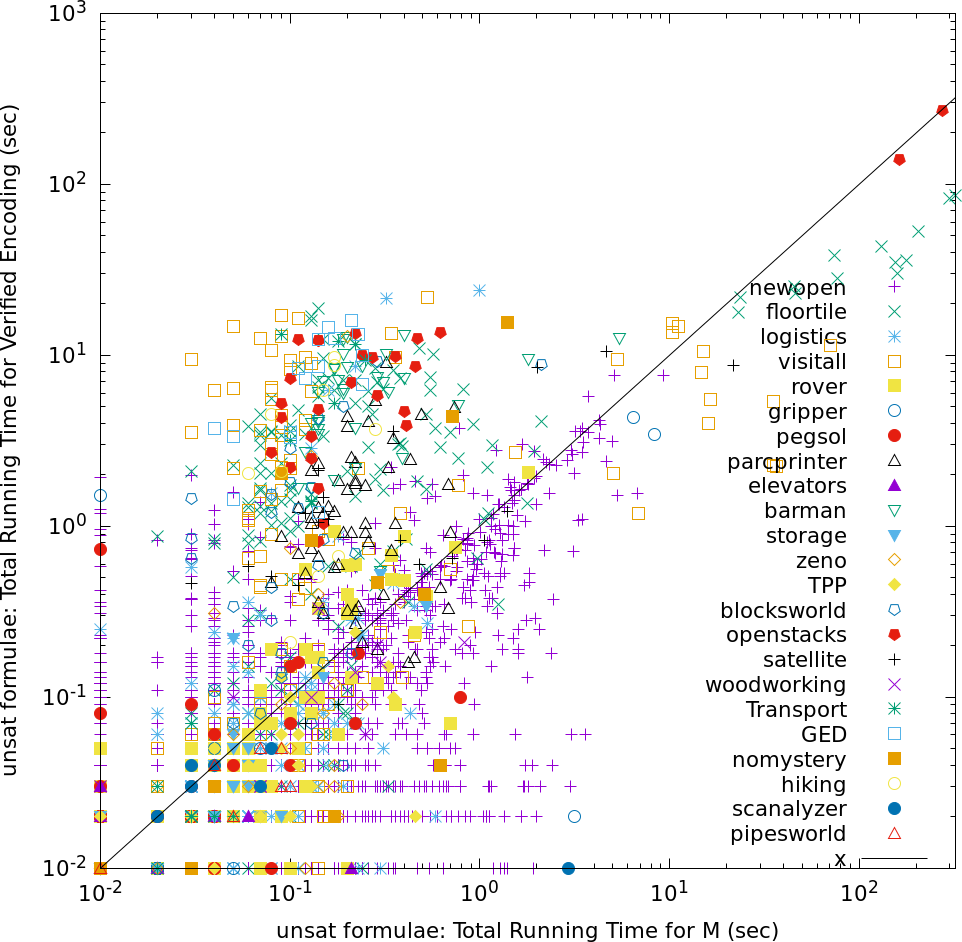}
\end{minipage}
\begin{minipage}[b]{0.24\textwidth}
\centering
        \includegraphics[width=1\textwidth,height=0.8\textwidth]{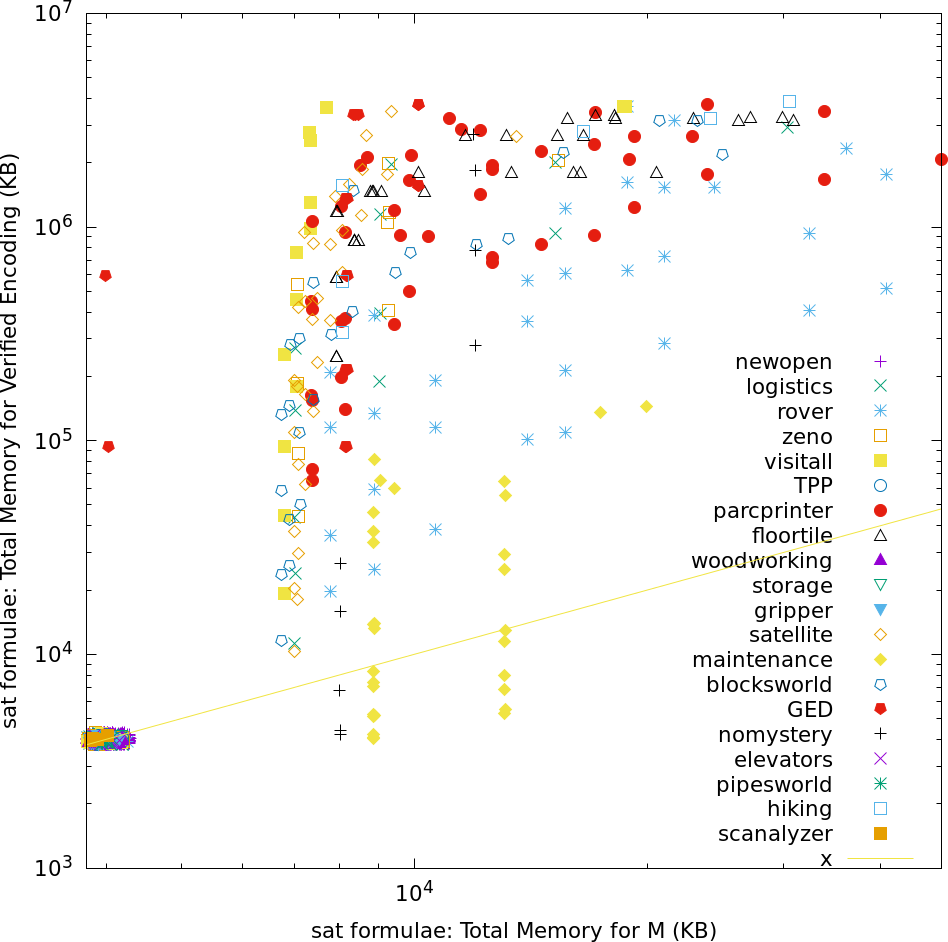}
\end{minipage}
\begin{minipage}[b]{0.24\textwidth}
\centering
        \includegraphics[width=1\textwidth,height=0.8\textwidth]{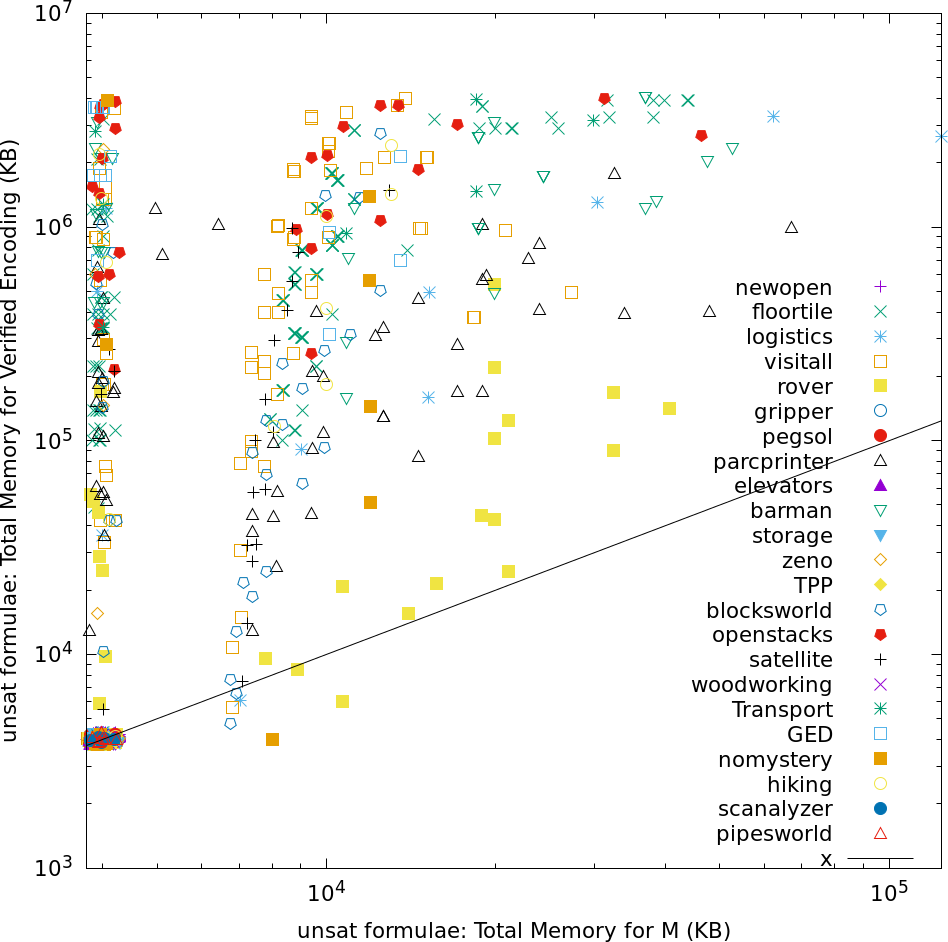}
\end{minipage}
     \caption{\label{figAppendix:timecompare}
Comparison of different performance indicators for our encoding and Madagascar's on satisfiable and on unsatisfiable formulae.
Different plots show the running times of Kissat, the number of clauses, the number of variables, the total running time of computing the encoding as well as solving the formulae, and the maximum memory needed for computing the encoding as well as solving the formulae.
}

\end{figure*}

\section{Appendix: Formalisation of Different Concepts in Isabelle/HOL}

\paragraph{Isabelle/HOL's syntax}
Isabelle's syntax is a variation of Standard ML combined with standard mathematical notation.
Function application is written infix, and functions can be curried, i.e.\ function $f$ applied to arguments $x_1~\ldots~x_n$ is written as $f~x_1~\ldots~x_n$ instead of the standard notation $f(x_1,~\ldots~,x_n)$.
Also, it supports lambda terms, e.g.\ $\lambda x\; y.\; x + y$ represents a function that takes two arguments and adds them.

We make use of the following predefined functions in Isabelle/HOL:\begin{enumerate*} \item \isabellefun{map} is the high-order function that returns a list after applying a function to each element of a given list, \item \isabellefun{fst} and \isabellefun{snd} are functions that extract the first and second members of a pair, \item \isabellefun{distinct} is predicate meaning that elements of a list are distinct, \item \isabellefun{filter} is a high-order function that removes members that do not satisfy a given predicate from a list, \item
\emph{[]} stands for an empty list, $\#$ stands for the list append operation, \item \isabellefun{set} takes a list and returns its set of elements, \item \isabellefun{@} concatenates two lists, \item \isabellefun{foldr} applies a function to elements of a given list, while accumulating the results, \item \isabellefun{zip} takes two lists of equal length and returns a list of equal length s.t.\ position $i$ in the output list is the pair of elements at position $i$ in the two given lists, \item for mappings $f$ and $g$, $f \subseteq_m g$ means that any argument mapped to a value by $f$ is mapped to the same value by $g$, \item \isabellefun{map\_of} takes a list of pairs and returns a function that corresponds to that list, \item $\circ$ denotes function  composition, \item  \emph{{\isacharplus}{\isacharplus}} takes two maps and merges them, with precedence to its right argument.
\end{enumerate*}

\subsection{Formalising Basic Definitions in Isabelle/HOL}
\begin{IsabelleSnippet}[label=isa:FDAST]{FD-AST and DIMACS-CNF}
type_synonym name = string
type_synonym ast_variable = name \<times> nat option \<times> name listing
type_synonym ast_effect = 
      ast_ass list \<times> nat \<times> nat option \<times> nat
type_synonym ast_operator =
      name \<times> ast_ass list \<times> ast_effect list \<times> nat
type_synonym ast_initial_state = nat list
type_synonym ast_goal = ast_ass list
type_synonym ast_problem = 
      ast_var_sec \<times> ast_initial_state \<times> ast_goal \<times> ast_op_sec
definition wf_partial_state ps \<equiv> distinct (map fst ps) 
           \<and> (\<forall>(x,v) \<in> set ps. x < numVars \<and> v < numVals x)
fun wf_operator :: ast_operator \<Rightarrow> bool 
  where wf_operator (name, pres, effs, cost) =
    wf_partial_state pres 
    \<and> distinct (map (\<lambda>(epres, v, l, m). v) effs) 
    \<and> (\<forall>(epres,x,vp,v) \<in> set effs. 
        wf_partial_state epres 
    \<and> x < numVars \<and> v < numVals x  
    \<and> (case vp of None \<Rightarrow> True | Some v \<Rightarrow> v<numVals x))
definition well_formed \<equiv> 
      length astI = numVars
    \<and> (\<forall>x<numVars. astI!x < numVals x)
    \<and> wf_partial_state astG
    \<and> (distinct (map fst ast\<delta>))
    \<and> (\<forall>\<pi> \<in> set ast\<delta>. wf_operator \<pi>)
definition valid_states :: state set 
  where valid_states \<equiv> {s. dom s = Dom \<and> (\<forall>x \<in> Dom. the (s x) \<in> range_of_var x)}
definition subsuming :: pstate \<Rightarrow> state set
  where subsuming partial \<equiv> {s \<in> valid_states. partial \<subseteq>\<^sub>m s}
definition eff_enabled :: state \<Rightarrow> ast_effect \<Rightarrow> bool 
  where eff_enabled s \<equiv> \<lambda>(epres,_,_,_). s \<in> subsuming (map_of epres)
definition enabled :: name \<Rightarrow> state \<Rightarrow> bool
  where enabled name s \<equiv> case lookup_operator name of
      Some (_,pres,effs,_) \<Rightarrow> s \<in> subsuming (map_of pres)
        \<and> s \<in> subsuming (map_of (implicit_pres effs))
    | None \<Rightarrow> False
definition execute :: name \<Rightarrow> state \<Rightarrow> state 
  where execute name s \<equiv> 
    case lookup_operator name of
      Some (_,_,effs,_) \<Rightarrow>
        s ++ map_of (map (\<lambda>(_,x,_,v). (x,v)) (filter (eff_enabled s) effs))
    | None \<Rightarrow> undefined
fun path_to 
  where path_to s [] s' \<longleftrightarrow> s'=s
  | path_to s (\<pi>#\<pi>s) s' \<longleftrightarrow> 
    enabled \<pi> s \<and> path_to (execute \<pi> s) \<pi>s s'
definition valid_plan :: plan \<Rightarrow> bool 
  where valid_plan \<pi>s \<equiv> \<exists>s' \<in> G. path_to I \<pi>s s'
\end{IsabelleSnippet}

To formally verify the encoding, we formalise the definitions of planning problems, SAT formulae, and their semantics in the language of Isabelle/HOL.
Referring to Listing~\ref{isa:FDAST}, we now review the formalisation of the definitions of FD-AST and DIMACS-CNF to give the reader a feel of this process and because the main theorems are stated in terms of them.
However, in the rest of the paper we only show the formalisation of definition or a concept when we think modelling it in Isabelle/HOL's language or as a functional program is interesting.

\paragraph{Abstract Syntax Trees}

Elements of the syntax tree are \emph{type synonyms}, i.e. abbreviations for types.
The most basic concept is that of a name, which is defined as an Isabelle/HOL string (line 1).
A variables, effects, operators and problems are is modelled as tuples with the obvious members (lines 2-8).
Note: \emph{nat option} models an optional natural number value.
Generally, an \emph{$\alpha$ option} is an algebraic data type consisting of the union between two types: one to represent no value and one that represents existence of a value of type $\alpha$.
For variables, we use it to express that a variable has an optional axiom layer.
An assignment (\isabellefun{ast\_ass}) is a pair, representing the variable and the assigned value.
For a problem, the variable and operator sections are lists of variables and operators, respectively.
The initial state is a list of natural numbers whose length has to be equal to the number of state variables, and the goal is a list  of \isabellefun{ast\_ass}.
We define functions to access the state variables, initial state, the goal and the set of actions of a given problem AST, and refer to them as \emph{astD}, \emph{astI}, \emph{astG} and \emph{ast$\delta$}, respectively.

The AST of DIMACS-CNF is a list of lists of integers, where each list represents a line in the DIMACS file, i.e.\ a clause, and each integer represents a literal.
The AST of a model for a DIMACS-CNF AST is a list of integers.

\paragraph{Well-Formedness}
Definitions of well-formedness of states, operators and problems are rather straightforward formalisations of Definition~\ref{def:FDASTwf}~(lines 13-26).
We note two parts of the Isabelle's syntax: \begin{enumerate*}\item the \isabellefun{case-of} statement in \isabellefun{wf\_operator} is a switch between the different constructors for a data type, and cases are separated with \isabellefun{|}, \item the predicate \isabellefun{well\_formed} takes an implicit argument \isabellefun{P} standing for the problem, but it is fixed outside the definition using a namespace like mechanism referred to as \emph{locale}.\end{enumerate*}

\paragraph{Semantics}

To formalise the semantics we define predicates and functions that describe action execution.
We first define two functions that, w.r.t.\ the problem, return the set of valid states and the set of valid states subsuming a given state (lines 27-31).
Based on that, Definition~\ref{def:FDASTex} is formalised (lines 31-43), and we recursively define a path between two states (line 44), and, lastly, define a predicate (line 48) characterising plans.

The semantics of CNF-DIMACS formulae and their models are formalised on lines 50-53.

\subsection{Formalising FDR and Encoding FD-AST}
\spacex{definition is_valid_operator_sas_plus
  :: ('variable, 'domain) sas_plus_problem  \<Rightarrow> ('variable, 'domain) sas_plus_operator \<Rightarrow> bool
  where is_valid_operator_sas_plus \<Psi> op \<equiv> let 
      pre = precondition_of op
      ; eff = effect_of op
      ; vs = variables_of \<Psi>
      ; D = range_of \<Psi>
    in list_all (\<lambda>(v, a). ListMem v vs) pre
      \<and> list_all (\<lambda>(v, a). (D v \<noteq> None) \<and> ListMem a (the (D v))) pre 
      \<and> list_all (\<lambda>(v, a). ListMem v vs) eff
      \<and> list_all (\<lambda>(v, a). (D v \<noteq> None) \<and> ListMem a (the (D v))) eff
      \<and> list_all (\<lambda>(v, a). list_all (\<lambda>(v', a'). v \<noteq> v' \<or> a = a') pre) pre
      \<and> list_all (\<lambda>(v, a). list_all (\<lambda>(v', a'). v \<noteq> v' \<or> a = a') eff) eff

definition is_valid_problem_sas_plus \<Psi> 
  \<equiv> let ops = operators_of \<Psi>
      ; vs = variables_of \<Psi>
      ; I = initial_of \<Psi>
      ; G = goal_of \<Psi>
      ; D = range_of \<Psi>
    in list_all (\<lambda>v. D v \<noteq> None) vs
    \<and> list_all (is_valid_operator_sas_plus \<Psi>) ops 
    \<and> (\<forall>v. I v \<noteq> None \<longleftrightarrow> ListMem v vs) 
    \<and> (\<forall>v. I v \<noteq> None \<longrightarrow> ListMem (the (I v)) (the (D v)))
    \<and> (\<forall>v. G v \<noteq> None \<longrightarrow> ListMem v (variables_of \<Psi>))
    \<and> (\<forall>v. G v \<noteq> None \<longrightarrow> ListMem (the (G v)) (the (D v)))}{}

\begin{IsabelleSnippet}[label=isa:FDR]{FDR problems and semantics}
type_synonym ('var, 'dom) state = 'var \<rightharpoonup> 'dom
type_synonym ('var, 'dom) assignment = 'var \<times> 'dom
record  ('var, 'dom) sas_plus_operator = 
  precondition_of :: ('var, 'dom) assignment list 
  effect_of :: ('var, 'dom) assignment list 
record  ('var, 'dom) sas_plus_problem =
  variables_of :: 'var list
  operators_of :: ('var, 'dom) sas_plus_operator list
  initial_of :: ('var, 'dom) state
  goal_of :: ('var, 'dom) state
  range_of :: 'var \<rightharpoonup> 'dom list
fun execute_serial_plan_sas_plus
  where execute_serial_plan_sas_plus s [] = s
  | execute_serial_plan_sas_plus s (op # ops) 
    = (if is_operator_applicable_in s op 
    then execute_serial_plan_sas_plus (execute_operator_sas_plus s op) ops
    else s) 
fun rem_effect_implicit_pres:: ast_effect \<Rightarrow> ast_effect where
  rem_effect_implicit_pres (preconds, v, implicit_pre, eff) = (preconds, v, None, eff) 
definition implicit_pres :: ast_effect list \<Rightarrow> ast_ass list 
  where implicit_pres effs \<equiv> 
    map (\<lambda>(_,v,vpre,_). (v,the vpre))
        (filter (\<lambda>(_,_,vpre,_). vpre\<noteq>None) effs)
fun rem_implicit_pres :: ast_operator \<Rightarrow> ast_operator where
  rem_implicit_pres (name, preconds, effects, cost) =
      (name, (implicit_pres effects) @ preconds, map rem_effect_implicit_pres effects, cost)
fun rem_implicit_pres_ops :: ast_problem \<Rightarrow> ast_problem where
  rem_implicit_pres_ops (vars, init, goal, ops) = (vars, init, goal, map rem_implicit_pres ops)
definition abs_ast_goal :: nat_sas_plus_state 
  where abs_ast_goal \<equiv> map_of astG
definition abs_ast_initial_state :: nat_sas_plus_state 
  where abs_ast_initial_state \<equiv> map_of (zip [0..<length astI] astI)
definition abs_range_map :: (nat \<rightharpoonup> nat list)
  where abs_range_map \<equiv> 
        map_of (zip abs_ast_variable_section 
                    (map get_range astDom))
definition abs_ast_operator :: ast_operator \<Rightarrow> nat_sas_plus_operator
  where abs_ast_operator \<equiv> \<lambda>(name, preconditions, effects, cost). 
        \<lparr> precondition_of = preconditions, 
          effect_of = [(v, x). (_, v, _, x) \<leftarrow> effects] \<rparr>
definition decode_abs_plan ops \<equiv> map (fst o the o lookup_action) (rem_condless_ops I ops)
\end{IsabelleSnippet}

Concepts from Definition~\ref{def:FDR} are formalised in Listing~\ref{isa:FDR}.
A \emph{state} is formalised as a \emph{mapping} from a variable type to a domain type.
An \emph{assignment} is formalised as a variable-value pair.
\isabellefun{'var} and \isabellefun{'dom} are polymorphic type parameters that can be instantiated.
These type variables are instantiated with the type \emph{nat} when translating from FD-AST to FDR.
An operator and an FDR problem are formalised as records.

The most interesting part is modelling the encoding and decoding as functional programs in Isabelle/HOL's language.
Encoding an FD-AST operator is done in multiple stages to make verifying the encoding more modular.
First, the effect preconditions are removed and added into the operator's preconditions.
This first step is further modularly implemented using \begin{enumerate*}\item \isabellefun{rem\_implicit\_pres\_ops}, which removes the effect preconditions from one given effect, and replaces it with \isabellefun{None}, \item \isabellefun{implicit\_pres} which extracts the implicit preconditions from a given effect, \item \isabellefun{rem\_implicit\_pres} which applies the previous two functions to an operator, thus lifting \isabellefun{rem\_implicit\_pres\_ops} to a list of effects using the higher-order function \isabellefun{map}, and lastly \item \isabellefun{rem\_implicit\_pres\_ops} which lifts the last function to an FD-AST problem by applying it to the list of operators in the problem. \end{enumerate*}.

The next step is translating an FD-AST resulting from applying \isabellefun{rem\_implicit\_pres\_ops}, whose operators have no effect preconditions, to FDR.
The variables in the FDR are the list of natural numbers below the number of variables in the FD-AST.
Since the FD-AST initial state is a list of values, the initial state is translated by just adding to every one of those values the variable to which it is assigned.
Constructing the FDR range map, which maps every variable to a list of natural numbers, is done using the function \isabellefun{abs\_range\_map}.
It is done by applying \isabellefun{zip} to two arguments, the first of which is the list of variables in the FD-AST, computed by \emph{abs\_ast\_variable\_section}.
The second argument is more involved.
The function \isabellefun{get\_range} computes the list of assignments a variable can take, given one entry from the variables section. 
The second argument to \isabellefun{zip} is thus a list of lists of natural numbers, each of which representing a the assignment of a variable.
The operators are translated in a straightforward way: the preconditions are taken as they are, and the effects are extracted by mapping a lambda function that, given an effect, extracts the affected variable and its new value as a pair.

The way we modularly divide the encoding of FD-AST problems into FDR problems made the proof of the first statement of Theorem~\ref{thm:asttofdr} more modular, and thus manageable.
A detail in the formal proof, which would be glossed over in a pen-and-paper treatment, is to show that translating enocding a well-formed FD-AST will result into a valid FDR, which is necessary for using the different theorems about FDR problems.
Proving that depends on the assumption that the AST actions have no conditional effects.
The second theorem that we prove about this translation is that it preserves completeness.

Modelling the decoding function is relatively straightforward and is done using \isabellefun{decode\_abs\_plan}.
Firstly note that $\circ$ denotes function composition.
The function \isabellefun{decode\_abs\_plan} firstly removes the operators whose preconditions are not statisfied when the operator sequence is executed at the initial state.
It then applies three steps to each of the operators in the given FDR plan: first it looks up the equivalent FD-AST operator, then it checks if the lookup operation was successful using the function \isabellefun{the}, and lastly it extracts the name of operator using \isabellefun{fst}, which returns the first element in a given tuple.

\mohammad{Add a note that we implemented $\choice$ as a list lookup in Isabelle}

\mohammad{Use the encoding and decoding envs}
\mohammad{Check paper RD.pdf to be shorten your pen-and-paper encoding and defs the encoding and decoding envs}

\subsection{Formalising the Encoding of STRIPS Problems into SAT}

Listing \ref{isa:SATPlan-encoding} shows parts of our encoding formalisation.
SATPlan variables are data types with two distinct constructors \texttt{State} and \texttt{Operator} storing time index and a number identifying the operator (line 1).
Lines 2-22 show the operator encoding. 
Importance was given to construct CNF formulas immediately to facilitate the proof that the encoding is in CNF form.
We use folds to construct the conjunctions with base value $\lnot\bot$.

\begin{IsabelleSnippet}[label=isa:SATPlan-encoding,float=h]{SATPlan encoding}
datatype sat_plan_variable = State nat nat | Operator nat nat  
definition  encode_all_operator_preconditions :: "'var strips_problem \<Rightarrow> 'var strips_operator list \<Rightarrow> nat \<Rightarrow> sat_plan_variable formula"
where "encode_all_operator_preconditions \<Pi> ops t \<equiv> 
  let l = List.product [0..<t] ops 
  in foldr (\<^bold>\<and>) (map (\<lambda>(t, op). encode_operator_precondition \<Pi> t op) l) (\<^bold>\<not>\<bottom>)"  
definition encode_all_operator_effects :: "'var strips_problem \<Rightarrow> 'var strips_operator list \<Rightarrow> nat \<Rightarrow> sat_plan_variable formula"
where "encode_all_operator_effects \<Pi> ops t \<equiv> 
  let l = List.product [0..<t] ops
  in foldr (\<^bold>\<and>) (map (\<lambda>(t, op). encode_operator_effect \<Pi> t op) l) (\<^bold>\<not>\<bottom>)" 
definition encode_operators :: "'var strips_problem \<Rightarrow> nat \<Rightarrow> sat_plan_variable formula"
where "encode_operators \<Pi> t \<equiv> 
  let ops = operators_of \<Pi> 
  in encode_all_operator_preconditions \<Pi> ops t \<^bold>\<and> encode_all_operator_effects \<Pi> ops t"   
\end{IsabelleSnippet}

\subsection{Formalising the Final Correctness Theorem}

\begin{IsabelleSnippet}[label=isa:finamThm]{Overall Encoding Correctness}
definition dimacs_model :: int list \<Rightarrow> int list list \<Rightarrow> bool 
  where dimacs_model ls cs \<equiv> (\<forall>c \<in> set cs. 
    (\<exists>l \<in> set ls. l \<in> set c)) 
      \<and> distinct (map dimacs_lit_to_var ls)
lemma encode_sound:
  \<lbrakk>valid_plan P \<pi>s; length \<pi>s \<le> h; encode h P = Inl cnf_formula\<rbrakk> \<Longrightarrow> 
         (\<exists>dimacs_M. dimacs_model dimacs_M cnf_formula)
lemma encode_complete:
  encode h P = Inr err \<Longrightarrow> 
    \<not>(well_formed P \<and> (\<forall>op \<in> set (ast\<delta> P). consistent_pres_op op) \<and>
    (\<forall>op \<in> set (ast\<delta> P). is_standard_operator op))
lemma decode_sound:
  decode M h P = Inl plan \<Longrightarrow> valid_plan P plan
lemma decode_complete:
  decode M h P = Inr err \<Longrightarrow>
          \<not> (well_formed P \<and> 
            (\<forall>op \<in> set (ast\<delta> P). consistent_pres_op op) \<and>
            (\<forall>\<pi> \<in> set (ast\<delta> P). is_standard_operator \<pi>) \<and> 
            dimacs_model M (SASP_to_DIMACS' h P))
\end{IsabelleSnippet}

In Isabelle/HOL formalisations of Theorem~\ref{thm:soundAndComplete} is shown in Listing~\ref{isa:finamThm} between lines (5-18).
We formalised the function \isabellefun{encode} in Isabelle/HOL's language s.t., given a horizon and FD-AST, it returns either \emph{Inl} and a SAT formula, if the encoding is successfully computed, or \emph{Inr} and an error message.
Similarly, \isabellefun{decode} returns either \emph{Inl} and a plan, if the decoding is successful, or \emph{Inr} and an error message otherwise.

\newpage

 \end{document}